\documentclass[twocolumn,letterpaper]{IEEEAerospaceCLS}  

\usepackage[]{graphicx}    
\usepackage{ amssymb }     
\usepackage{hyperref}
\usepackage{cite}
\usepackage{colortbl,hhline,xcolor}
\usepackage{gensymb}
\usepackage{tabularx}
\usepackage{xurl}

\newcommand{\ignore}[1]{}  
\usepackage{subcaption}

\begin{document}
\title{Battery-Swapping Multi-Agent System for Sustained Operation of Large Planetary Fleets}
\author{%
Ethan Holand$^\dag$, Jarrod Homer, Alex Storrer, Musheeera Khandeker, Ethan F. Muhlon, Maulik Patel, Ben-oni Vainqueur, \\ David Antaki, Naomi Cooke, Chloe Wilson, Bahram Shafai, Nathaniel Hanson,  Taşkın Padır\\ 
Northeastern University - Institute for Experiential Robotics\\
360 Huntington Ave,\\ 
Boston, MA 02115\\
\small\{holand.e, homer.j, storrer.a, khandaker.m, muhlon.e, patel.maul, vainqueur.b, \\
\small antaki.d, cooke.n, wilson.chl, b.shafai, hanson.n, t.padir\}@northeastern.edu
}

\maketitle
\thispagestyle{plain}
\pagestyle{plain}
\begin{abstract}
We propose a novel, heterogeneous multi-agent architecture that miniaturizes rovers by outsourcing power generation to a central hub. By delegating power generation and distribution functions to this hub, the size, weight, power, and cost (SWAP-C) per rover are reduced, enabling efficient fleet scaling. As these rovers conduct mission tasks around the terrain, the hub charges an array of replacement battery modules. When a rover requires charging, it returns to the hub to initiate an autonomous docking sequence and exits with a fully charged battery. This confers an advantage over direct charging methods, such as wireless or wired charging, by replenishing a rover in minutes as opposed to hours, increasing net rover uptime. 

This work shares an open-source platform developed to demonstrate battery swapping on unknown field terrain. We detail our design methodologies utilized for increasing system reliability, with a focus on optimization, robust mechanical design, and verification. Optimization of the system is discussed, including the design of passive guide rails through simulation-based optimization methods which increase the valid docking configuration space by 258\%. The full system was evaluated during integrated testing, where an average servicing time of 98 seconds was achieved on surfaces with a gradient up to 10\degree. We conclude by briefly proposing flight considerations for advancing the system toward a space-ready design. In sum, this prototype represents a proof of concept for autonomous docking and battery transfer on field terrain, advancing its Technology Readiness Level (TRL) from 1 to 3.
\end{abstract} 

\tableofcontents


\thanks{
\footnotesize$\dag$ Corresponding author.\\
\footnotesize Project code, documentation, and CAD models available online at \url{https://river-lab.github.io/BOOST}.\\
\footnotesize 979-8-3503-0462-6/24/$\$31.00$ \copyright2024 IEEE
}              

\begin{figure}
    \centering
    \includegraphics[width=1\linewidth]{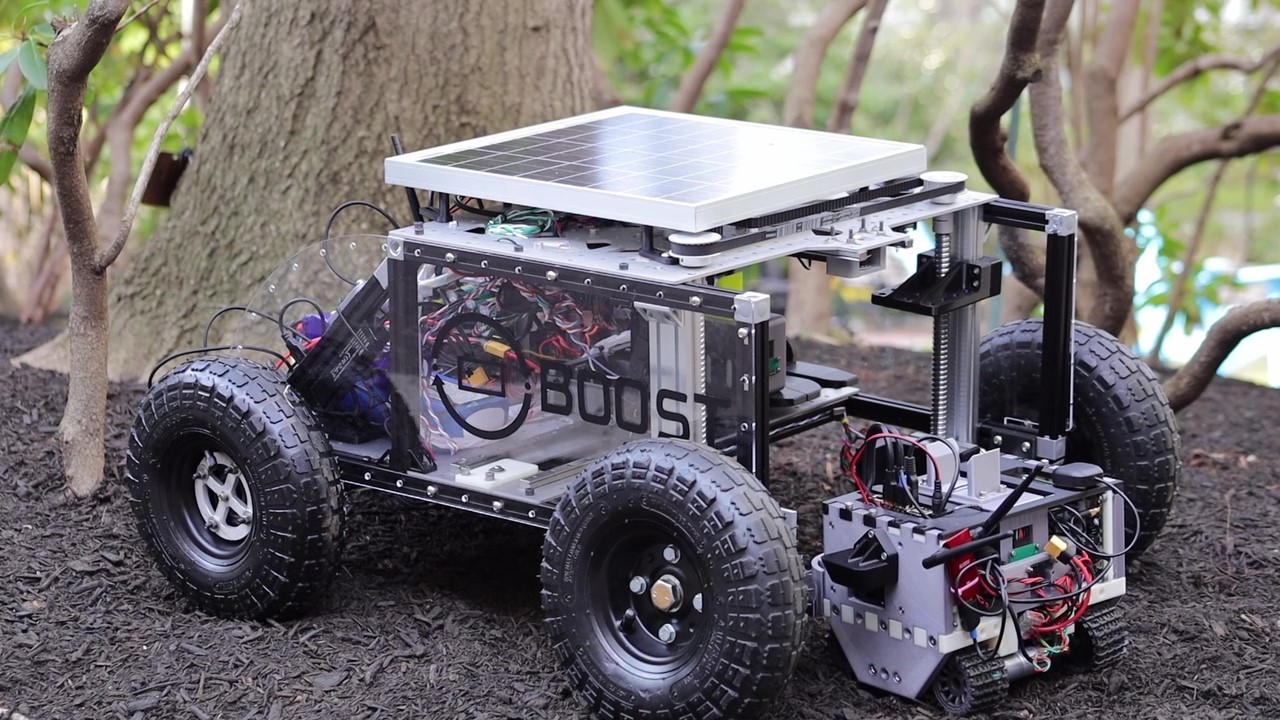}
    \caption{Field Testing of the prototype battery-swapping multi-agent rover architecture. The large rover, left, generates and distributes power to an internal cache of battery modules. The small rover, right, docks with the base to hot-swap its spent battery for one with full charge.}
    \label{fig:boostglamor}
\end{figure}

\section{Introduction}

In the near decade, NASA seeks to establish a permanent presence on the Moon through the Artemis program. During Phase 2, a network of surface habitats, rovers, and instruments are proposed to sustain the human return to the Moon \cite{artemis}. Both the generation and distribution of power are critical challenges toward achieving this; as stated within the Artemis Lunar Surface Technology Strategy, surface power technology for sustainable continuous power is a key priority of NASA's Space Technology Mission Directorate (STMD) \cite{artemisplan}. 

We suggest that portable and rechargeable battery modules offer a useful addition to a power distribution portfolio. Backup modules may be pre-charged at a base while field activities are being performed. When the rover battery power depletes, a fresh battery can quickly be exchanged, virtually eliminating the downtime associated with direct rover charging. In addition to powering rovers, battery modules may offer additional mission flexibility, such as the ability to power tools for astronaut field operations.

In this work, we present a framework that implements battery swapping to distribute power from a central generator to a fleet of rovers. This system consists of a base rover which serves as the centralized communication and power hub. Smaller field rovers are equipped with a battery terminal and mating geometry to aid in the docking process. An autonomous docking and swapping protocol is implemented to service rovers in under 2 minutes.

The core contributions of this work are:

\begin{itemize}
    \item Advancement of the Technology Readiness Level (TRL) of battery-swapping for field and space robotics. 
    \item Development and presentation of a low-cost, open-source battery-swapping test platform.
    \item Optimization-based design of the robots' physical structure to improve performance via rigid-body docking simulations.
\end{itemize}

\section{Background}

\subsection{Existing Power Generation Solutions}
Historically, planetary rovers have been directly outfitted with the power generation capacity to meet their needs. Two categories of power generation are used on existing rovers; photovoltaic solar arrays, as used in Spirit and Opportunity, or Radioisotope Thermoelectric Generators (RTG), as used in Curiosity and Perseverance \cite{Solar_rovers,curiosity,perserverence}. 

Solar cells utilize the photovoltaic effect to generate renewable energy. However, the need for direct solar exposure produces several system limitations. Direct sunlight may not always be available, such as at night, in permanently shadowed regions, or within craters, canyons, or lava tubes. Due to the inverse square law, solar panel size increases rapidly for equivalent power generation deeper into the solar system. Additionally, airborne regolith and dust can collect on solar cells, limiting their run time and efficiency.

RTG power generation takes advantage of the natural decay of radioactive material to provide a high-temperature source. With this, the temperature gradient in a p-n junction acts as a heat pump, generating energy. RTGs confer several advantages for interplanetary missions as they are completely encased and continuously generate power, regardless of distance from the sun. However, RTGs offer relatively low efficiency, only converting around 6.5–7\% of their power output to electrical energy \cite{rtgefficiency}. 

\begin{table*}
    \centering
    \begin{tabular}{|p{0.1\linewidth}|p{0.25\linewidth}|p{0.25\linewidth}|p{0.25\linewidth}|p{0.05\linewidth}|} \hline
         \rule{0pt}{2.2ex}\textbf{Rover Power Method } &  \textbf{Advantages} &  \textbf{Disadvantages} &  \textbf{Existing Work} & \textbf{TRL} \\ \hline\hline 
         \rule{0pt}{2.2ex}Onboard Solar&  Low-cost. Safe, as opposed to nuclear solutions. Relatively lightweight. &  Dust accumulation on panels greatly limits performance over time. 
Solar area grows quadratically as distance from the sun increases. Nighttime or shadowed operation is limited.&  Demonstrated on Sojourner, Spirit, Opportunity, and Ingenuity.& \cellcolor{green}High (9)\\ \hline 
         \rule{0pt}{2.2ex}Onboard RTG&  Small but reliable power for long mission lifespans. Environment agnostic. Doubles as a heat source.&  Extremely low electrical power output. Limited supply of Plutonium. Radioactive material challenges construction and poses launch dangers.&  Demonstrated on Curiosity \& Perseverance.& \cellcolor{green}High (9)\\ \hline 
          \rule{0pt}{2.2ex}Externally Generated, Tethered & High power transmission with low efficiency loss. No servicing downtime. Potential communication capabilities. & Entanglement risk with other rovers or environmental hazards. Drag forces impede vehicle travel. Limited range and maneuverability. & Tethered Power Systems for Lunar Mobility and Power Transmission (TYMPO), awarded 2021-2023 NASA STMD \cite{tether} & \cellcolor{yellow}Mid (4-6) \\ \hline 
         \rule{0pt}{2.2ex}Externally Generated, Wirelessly Charged&  No moving parts. Accommodates various battery sizes. Resistant to environmental conditions such as dust. &  Large robot downtime; half of the robot lifespan may be spent charging. Wireless transmission inefficiencies.&  2021 NASA Tipping Point awarded to Astrobotic, Bosch, UW, and WiBotic \cite{wireless1}. 2023 NASA SBIR awarded to Yank Technologies \cite{wireless2}& \cellcolor{orange}Mid (4-5)\\ \hline 
         \rule{0pt}{2.2ex}Externally Generated, Battery Swapping&  Virtually no system downtime. High power efficiency. Battery modules may be extended for astronaut tool use.&  Increased design complexity. Moving parts and electrical contacts risk dust buildup and mechanical failure. Increased mass and volume of hub. &  Existing research for earth-based EVs and drones. No known work for field robotic demonstrations.& \cellcolor{red}Low (1)\\ \hline
    \end{tabular}
    \caption{Comparison between onboard and external powering methods for rovers.}
    \label{tab:powercompare}
\end{table*}

\subsection{The Future of Power Generation}

A new class of power generation may be enabled via nuclear fission generators. NASA and the DOE's \textit{Kilopower} research effort has investigated the implementation of 1-10 kW fission reactors. In 2018, an experimental fission reactor successfully produced 5.5 kW of electrical power \cite{kilopower}. In 2022, NASA and the DOE awarded three \$5 million contracts for 40-kW class fission systems \cite{FissionAwards}. 

These metrics far outclass the generation of existing rovers. The RTGs on Curiosity and Perseverance produce 0.11 kW of electrical power and 2.2 kW of heat, some of which is used to heat the spacecraft, the remainder of which is rejected \cite{rtg}. The solar cells on Spirit and Opportunity produced 0.9 kW at their peak and 0.41 kW during their extended mission \cite{SpiritandOpportunityPower}. Thus, the demonstrated 5.5 kW generator could sustain about 6-12 similar rovers, and a future 40 kW generator could support multiple dozen.

Large multi-agent systems are desirable for both science and infrastructure. From a sensing perspective, a fleet of robots can collect a temporally synchronized set of data points, enabling deeper insights. For example, NASA's planned 2024 mission, CADRE, will demonstrate the capability of transmitting simultaneous ground-penetrating radar pulses to reconstruct 3D subsurface maps \cite{CADRE}. Large fleets may additionally prove critical in preparing lunar habitats for Artemis, leveraging in-situ resource utilization (ISRU) for habitat construction \cite{isru}.

These power systems offer a new paradigm for planetary surface operations that could more closely mimick the electrical grid on Earth. Here, power generation and consumption are separated, enabling efficiency gains on either end. Large, dedicated power generation systems may be designed to be hyper-efficient without the constraints of fitting on a rover's moving back. Further, rovers without power generation systems save on SWAP-C. Without this mass on board, rovers would expend less energy to travel the same distance.

\subsection{Power Transfer from Hub to Rover}
With a central power-generating hub, a large number of battery-powered rovers can be supported. The hub acts as a refueling station, enabling the rovers to refill when their power runs low. There are three primary avenues for electrical power transfer from the hub to a rover: wireless charging, wired charging, or physical transfer of power modules, henceforth referred to as battery swapping.

Wireless charging poses a low-risk power transfer method, as it is robust to dust ingress, may function even with misalignment, and requires no moving components. Wireless transmission additionally offers the flexibility to operate with rovers of various form factors given a standardized receiver. These advantages have prompted recent NASA Small Business Innovation Research (SBIR) and Tipping Point awards \cite{wireless1, wireless2}. Prototypes created from these grants successfully transmitted power in extreme environments, achieving 80-85\% transmission efficiency \cite{astroboticwireless}.

A primary limitation of traditional wired or wireless charging is the rover downtime during service. Using optimized charging protocols and lower charge rates to preserve battery health, a lithium ion battery of any capacity can be charged in approximately 2-3 hours \cite{ChargeProtocol, ChargeRate}. While novel approaches are being researched to enable fast charging with minimal battery degradation, the research conducted remains focused on EVs and currently has low TRL \cite{EVFastCharge}.

Conversely, the physical transfer of power modules enables rapid refueling at the cost of system complexity. By pre-charging a replacement battery while a rover is exploring, the rover can be replenished with a fully charged battery in minutes, about two orders of magnitude lower than wireless servicing \cite{ChargeRate}. This enables near-continuous coverage of a region with virtually no rover downtime, at the cost of rover dependence and additional system complexity for the battery management systems.

A comparison between onboard and shared power sources is presented in Table \ref{tab:powercompare}. The evaluated TRL of battery swapping is low relative to alternate solutions, motivating this work.

The majority of prior battery-swapping work has been concentrated on the use case of electric vehicles. A variety of models have been proposed to optimize the planning of such fleets \cite{batterymanagement1, batterymanagement2}. Some prototype hardware platforms for mobile robots have been developed, but those surveyed were designed for indoor home or laboratory environments \cite{batteryswap1, batteryswap2, batteryswap3, batteryswap4}. Existing work for field environments has been relegated to unmanned aerial vehicles (UAVs) \cite{uavswap, uavswap2}. No known works exist within an aerospace application. As a result, the focus of this paper is on developing a ground-based platform for autonomous docking and swapping on unknown terrain.


\section{Generalized Architecture}
\begin{figure*} 
\centering
    \includegraphics[width=\linewidth]{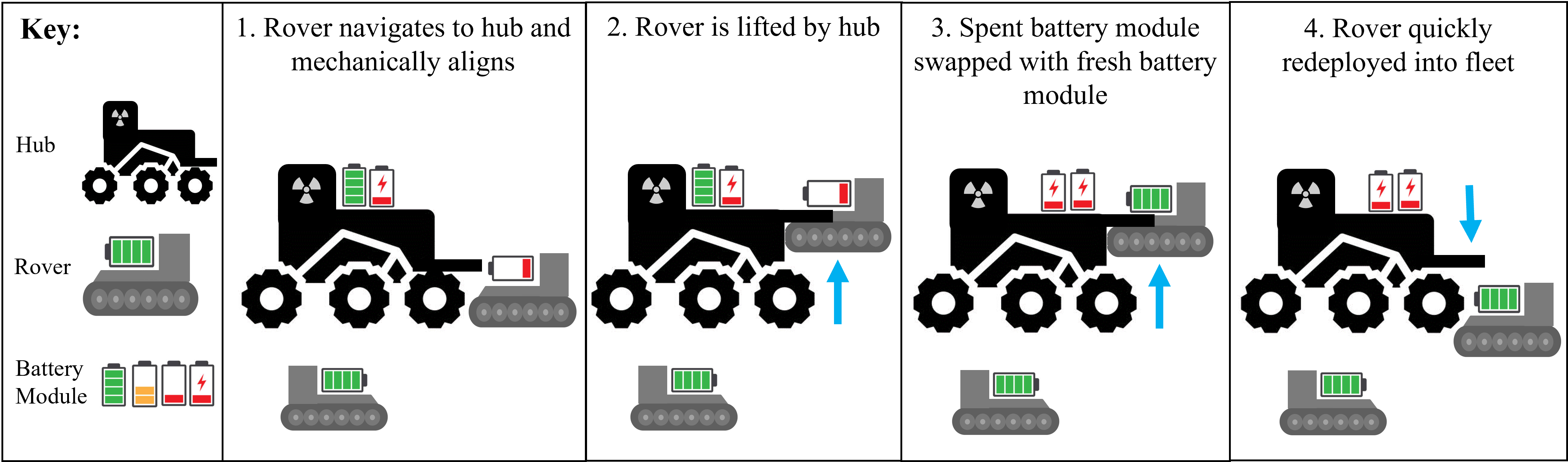}
    \caption{Battery swapping method represented schematically.}
    \label{SystemDiagram}
    \vspace{-1.0em}
\end{figure*}

The core parts comprising a battery-swapping architecture are rovers, outfitted with a battery terminal; hubs, which possess power generation capabilities, a battery cache, and swapping hardware; and standardized battery modules, designed for ease of removal and insertion. To achieve the battery transfer, a docking and swapping process is needed. Fig.~\ref{SystemDiagram} provides an initial overview of the battery swap architecture for further visualization of the system. 

\subsection{Rovers:}
Rovers comprise the fleet of robots tasked to navigate and perform work on a planetary surface. They may be specialized to carry payloads and instruments. Each contains a battery module, battery terminal, and the corresponding hardware required for docking.

\subsection{Power Generation Robot (Hub):}
One or more hubs, either mobile or stationary, may sustain a fleet of rovers. Hubs possess large and efficient power-generation sources, such as directed solar arrays, large RTGs, or fission reactors. Using an internal battery management system, they store the generated power in multiple battery modules. Each hub is outfitted with one or more docking ports necessary to conduct a battery swap. Where possible, components necessary for the battery swap should be relegated to the hub to minimize added mass and complexity per rover.

The hub may serve additional functions that further enable the miniaturization of each rover. For example, the hub may serve as a base link, hosting long-range antennas for communicating with orbiters or the deep space network. The hub may carry heavy computing power to process rovers' data and direct planning of the network. Finally, the hub may offer additional mission-specific support, such as the ability to process geological samples delivered by a rover.

The number of rovers a single hub can support, $n_r$, can be defined by:

\begin{equation}
n_r=\left\lfloor\frac{P_{gen}-P_h}{\max(Q_bV_b/C, P_m)}\right\rfloor
\label{PowerGeneration}
\end{equation}

where $P_{gen}$ is the power generation of the hub, $P_h$ is the power consumed by the hub during operation, $Q_b$ is the capacity of a battery module, $V_b$ is the voltage of a battery module, $C$ is the charge rate of the hub, and $P_m$ is the average power consumption per rover. The maximum of the denominator is taken to account for bottlenecks due to battery module charge rate or rover consumption rate, respectively.

If a specific mission requires a larger area of coverage, a network of hubs can be used to support a larger fleet. Several different considerations are relevant in determining the area of coverage of a multi-agent fleet including the number of hubs, rover battery consumption, and rover velocity. These relationships and hub network configurations are discussed in detail in Appendix \ref{sec:hubcoverage}.

\subsection{Docking}
Within the docking stage, an approaching rover is aligned to a known position relative to the hub. Mechanical alignment and visual perception may be used to aid the docking phase, enabling rovers to consistently reach a known position from a wide range of initial configurations.

\subsection{Swapping}
Once the rover is in a known position, the battery swap may be conducted. The following key steps are conducted in order to swap a battery module:
\begin{enumerate}
    \item Removal of discharged battery from rover
    \item Stowing of discharged battery to be recharged
    \item Selection of fully charged battery to be deposited
    \item Insertion of fully charged battery into rover
\end{enumerate}


\section{Prototype Implementation}
\subsection{Overall Considerations}
\begin{figure}
    \centering
    \includegraphics[width=1\linewidth]{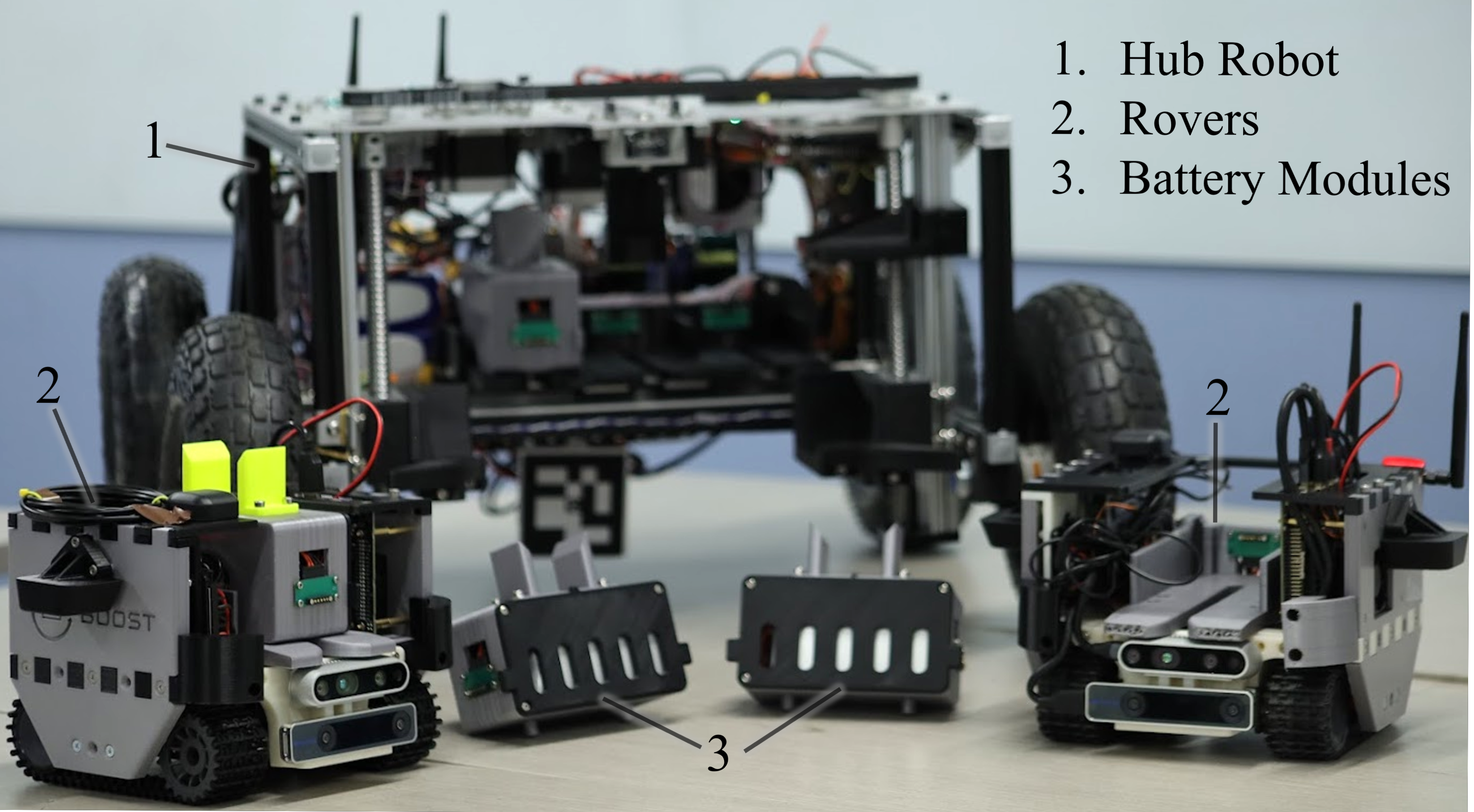}
    \caption{Final prototype with labeled components.}
    \label{fig:boost photo}
\end{figure}

Given the generalized architecture, a proof-of-concept system was created to evaluate the implementation of a battery-swapping system for field robots. For the prototype system, two rovers, one hub, and four battery modules were developed, as seen in Fig.~\ref{fig:boost photo} and Appendix \ref{sec:Renders}. A table summarizing characteristic performance parameters is presented in Appendix \ref{sec:physparams}. A fully autonomous docking and swapping protocol was implemented. The result is a baseline proof of concept that can be further refined into a higher TRL in future revisions.

The size of the rover is governed by the payload that the rover is expected to carry, which will vary for different applications. The size of the hub is a function of the rover size and number of battery modules. Although a direct payload was not implemented in this prototype, the rovers were sized off of a representative payload of a ground penetrating radar (GPR). Referencing RIMFAX, the GPR system on Perseverance, a rough mass (1.245 kg), volume (196 x 135 x 66 millimeters), and power (2.5-9.5W) were gathered as driving criteria \cite{rimfax}. Other sensors such as X-ray spectrometers, radiation detectors, cameras, and thermal emission spectrometers like those used on Opportunity, Spirit, and Curiosity, also meet this small form factor design requirement \cite{instruments}.

Second, we prioritized docking and swapping reliability, with the goal of being able to complete the entire docking and battery-swapping process autonomously. We aimed to accomplish this by designing large mechanical compensation and adding hardware and software verification methods in each step of the docking procedure.

Third, in our design process, we targeted ease of system realization. Some concessions were made to simplify development for an earth-based system: for example, space-proofing via dust shielding, thermal insulation, and radiation hardening were not implemented. A direct rover payload was not implemented other than the visual sensing suite. A solar panel was outfitted aboard the hub, but it was not connected during testing. Commercial-off-the-shelf parts were used throughout the prototype, such as the rover's tracked mobility system, which while not realistic to a space implementation, streamlined design time toward a proof of concept on Earth.

\begin{figure*} 
        \includegraphics[width=\linewidth]{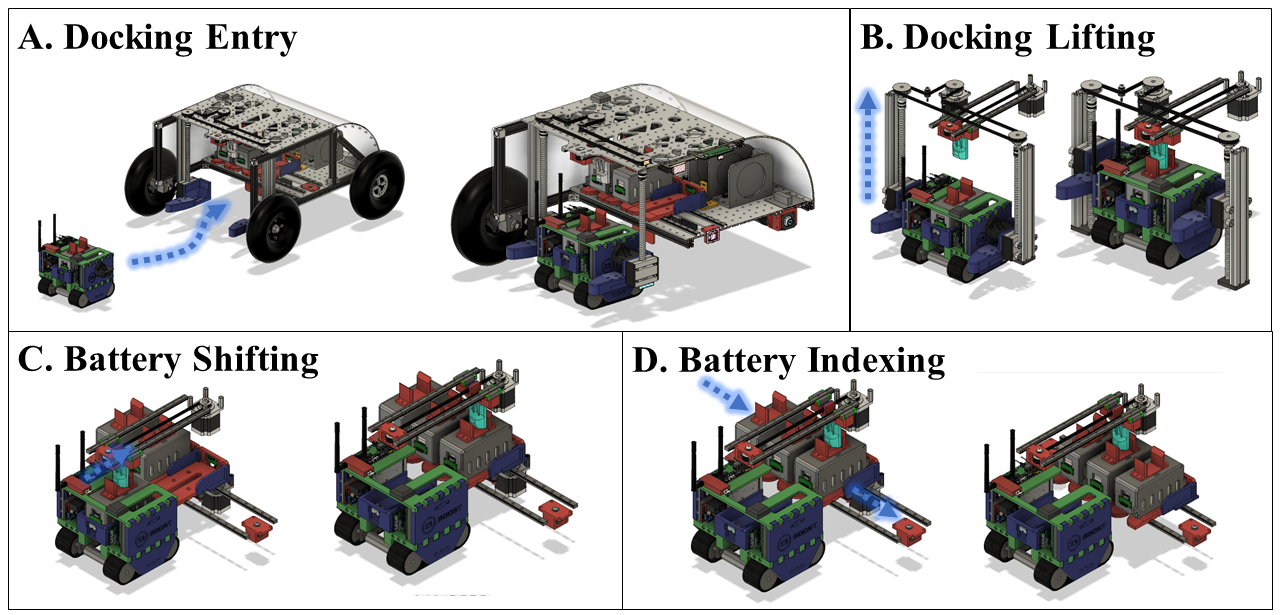}
        \caption{Step-by-step process diagram of docking and battery swap procedure. After step D, the procedure is reversed to complete a swap.}
        \label{dockingoverview}
\end{figure*}

The docking and battery swapping system implemented is described in the following subsections and shown schematically in Fig.~\ref{dockingoverview}.

\subsection{Battery Module}
In order to facilitate the mechanical transfer of a battery between a rover and a hub, a 6S-LiPo battery was packaged in a 3D-printed case with additional electronics to allow for charging and ease of transfer. The battery was chosen for its capacity to power a rover for two hours on a full charge. Mechanically, the battery module includes a tab on the top to engage with the battery swapping mechanism to transfer batteries between the hub and rover. The case features vents near the battery to ambiently dissipate heat. Both ends of the battery module are unique to either charge or discharge, so a hard stop was implemented on one side to prevent the battery from being plugged in inverted.

\begin{figure*}     
    \includegraphics[width=\linewidth]{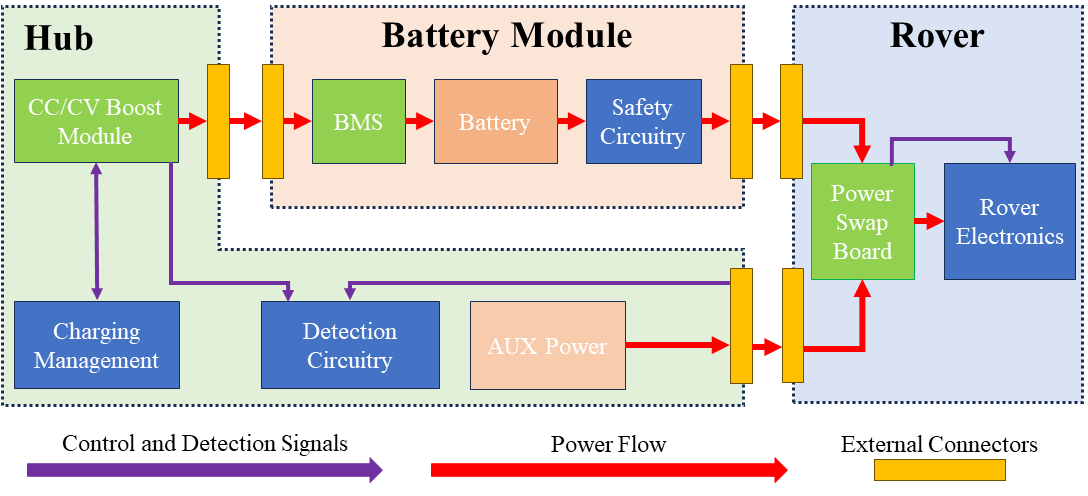}
    \caption{Block diagram representing critical electrical components for power transfer across a hub, battery module, and rover.}
    \label{PowerFlow}
\end{figure*}

Electrically, the battery module includes a battery management system (BMS) for balancing the six cells, a fuse to protect the module if too much current is drawn, and a relay that allows for the battery module to activate once continuity is detected via custom printed circuit boards (PCBs). An overview of the systems in place for power management of battery modules and docking and swapping verification components is shown in Fig.~\ref{PowerFlow}.

\subsection{Docking Procedure}
The docking process prepares a rover for a battery swap by aligning it to a known position relative to the hub, as shown in Fig.~\ref{dockingoverview} A and B. This is achieved using a two-step process. First is docking \textit{entry}, in which the rover is coarsely aligned to position by entering the docking port. Next is docking \textit{lifting}, in which notched lifting arms elevate the rover off unknown terrain. 

\subsubsection{Docking Entry}
During docking entry, the rover may approach from a variety of poses. A combination of machine vision and mechanical alignment is utilized to compensate for a wide range of approach trajectories and relative orientations between the hub and rover.

The rover's software platform runs on Ubuntu with ROS2 and Jetpack, using the NAV2 Stack for autonomous navigation \cite{ROS2, NAV2}. ARuCO tags are utilized as fiducials, which are set as navigational waypoints once detected with OpenCV\cite{garrido2014automatic, OpenCV}.

The rovers utilize an Extended Kalman Filter (EKF) within the ROS2 NAV2 Stack \cite{EKF}. This sensor data is gathered from the following sources for the following uses:
\begin{itemize}
    \item Intel D435 Depth Camera (environmental map)
        \item Pointcloud, RGB Camera, Depth Image (point cloud converted to laser scan to reduce computation)
    \item Intel T265 Camera (localization and motion tracking) with internal IMU/Gyro, Angular/Linear Velocities, and two fish eye lenses
    \item Custom PCB / STM32 with IMU/Gyro, GPS, Battery status, Encoder data
\end{itemize}

Once a rover commences its docking sequence, a proportional-integral controller is employed to minimize the angular, positional, and distance errors with the fiducial while docking entry is conducted.

\begin{figure*} 
    \includegraphics[width=\linewidth]{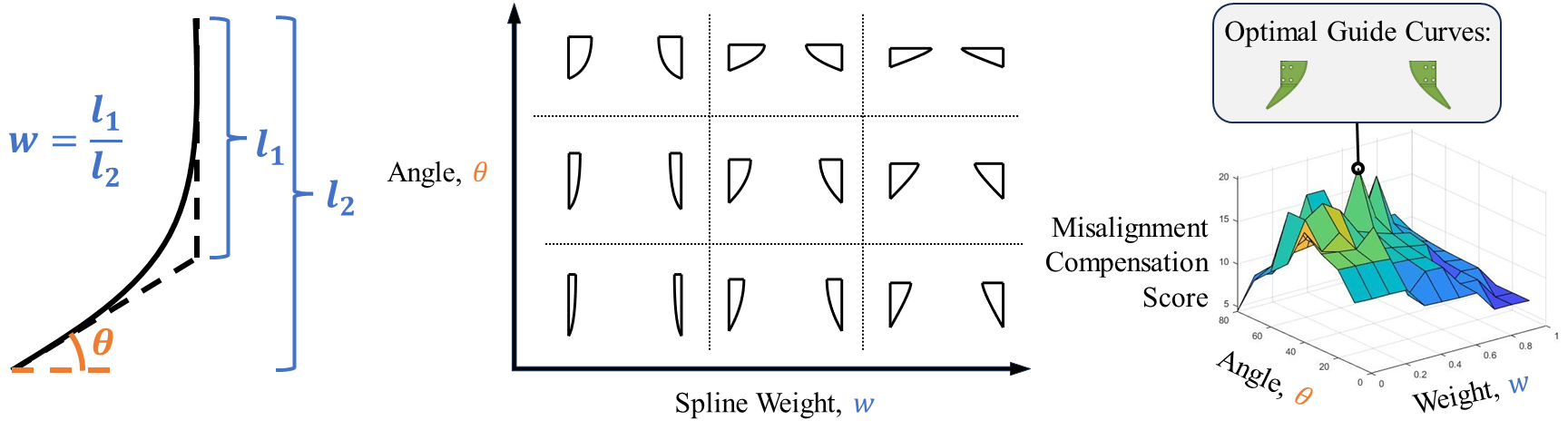}
    \caption{Bezier entry guide curve grid search to optimize bumper design. Spline angle $\theta$ represents the angle between the horizontal and the lower tangent. Spline weight $w$ controls the relative position of the tangency point. The optimally generated profile is shown on the right.}
    \label{EntryCurve}
    \vspace{-1.0em}
\end{figure*}

The visual navigation is compensated by 3D-printed entry guide bumpers, shown in Fig.~\ref{EntryCurve}, which physically guide the rover into the docking port. The bumper geometry was optimized using rigid-body docking simulations conducted in Matlab using the Simscape Multibody plugin. The simulation enabled the rapid testing of bumper geometries by varying the rover's starting orientation to see if this would lead to a successful docking attempt. The bumper curve was modeled as a quadratic Bézier curve, simplified to two parameters \cite{baydas2019defining}. Using an iterative grid search, the Matlab script evaluated 111 spline configurations with spline tangencies and weights ranging from $0-90\degree$ and $0.0-1.0$, respectively. For each configuration, a binary search was conducted to evaluate the maximum angular and axial compensations; the L2 norm of both was taken to provide a net compensation score. From this, the highest-scoring curve was selected as the optimal guide curve. The entry curve parameters and grid search are shown in Fig.~\ref{EntryCurve}.

\begin{figure*} 
    \includegraphics[width=\linewidth]{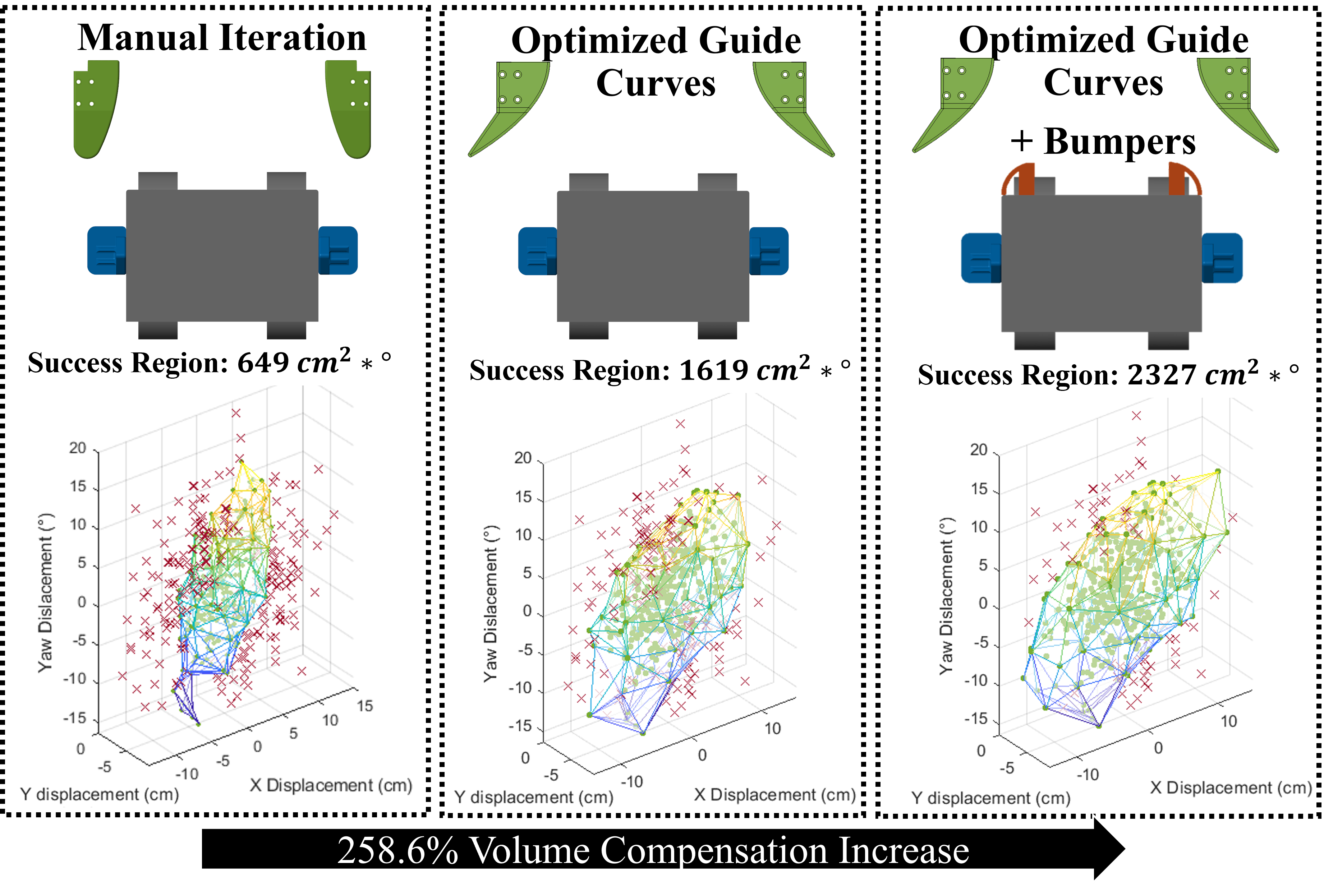}
    \caption{Simulation results showing success region volume for original entry guide curve (left), optimized Bézier guide curves (center), and optimized guide curves with bumpers (right).}
    \label{OptimizedHull}
\end{figure*}

To evaluate combined axial and rotational displacements, a robustness metric was created using the volume of the configuration space. First, Monte-Carlo simulations were conducted by sampling a Gaussian distribution containing a large number ($n>500$) of initial poses, representing random displacements and misalignments between the mating hub and rover components. For each configuration, a simple pass-fail docking test was simulated and recorded. Generating a 3D plot from the results of these simulations reveals a central configuration space within which any orientation of the rover can be mechanically compensated by the hub. A convex hull was wrapped around this success region using the quickhull algorithm \cite{convhull}. By taking the volume of this hull, a robustness metric was created for direct comparison between major system iterations. This further motivated the design choice to implement bumpers to the front of the rover as shown in Fig \ref{OptimizedHull} (right) to further increase the robustness. Using this simulation approach, the success region volume increased by 258\% from the initial to the optimized iteration.

After the rover is successfully guided into the docking port, it is mechanically stopped by contact between the rover's side tabs and the hub's hardstops. The rover's drive system brakes upon sensing that a fiducial marker is in range and the forward acceleration has ceased. The rover is now in position for the docking lifting step.

\subsubsection{Docking Lifting}

Once the rover has successfully entered the docking port of the hub, the hub's lifting arms raise the rover by a pair of lifting tabs. Via the effects of gravity, the sloped profiles of the arm guide the tabs into a lower resting position. In addition to this gravitational alignment, magnets in the arms snap the rover into a final resting position. 

Within this position, custom electrical contacts between the hub and rover are engaged. These are first utilized to conduct a continuity check in order to determine whether the docking process was successful. Additionally, these contacts transfer auxiliary power to the rover during a swap, ensuring that the rover remains powered when the spent battery is removed, which may be critical to maintaining communication, system monitoring, and thermal conditions in a space implementation.

Lifting is stopped when the rover is level with the hub's battery cache. At this height, a bi-directional pusher nests within two tabs protruding from the battery module. These tabs are offset to allow $\pm5$mm of positional tolerance on the pusher. This primes the system for the battery swap procedure without requiring any additional actuation. 

\subsection{Battery Swap Procedure}
The battery swap procedure consists of two key actions: \textit{indexing}, in which the battery cache is translated laterally to select a battery terminal, and \textit{shifting}, in which battery modules are transferred between the rover and hub. These are depicted in Fig.~\ref{dockingoverview} C and D. A full swap sequence occurs by indexing to an empty terminal, shifting the rover's battery module to this terminal, indexing to a fresh battery, and shifting it back into the rover. 

Several design considerations were implemented to simplify the battery swap process, lower risk, and increase allowable system tolerance, as follows. 

The lifting arms are programmed to slightly elevate the rover when transferring a module to the hub and slightly lower the rover when receiving a module from the hub. This offset mitigates the risk of a battery module jamming at the lip of the terminals and allows tolerance in the lift height position.

\begin{figure*}
    \centering
    \includegraphics[width=\linewidth]{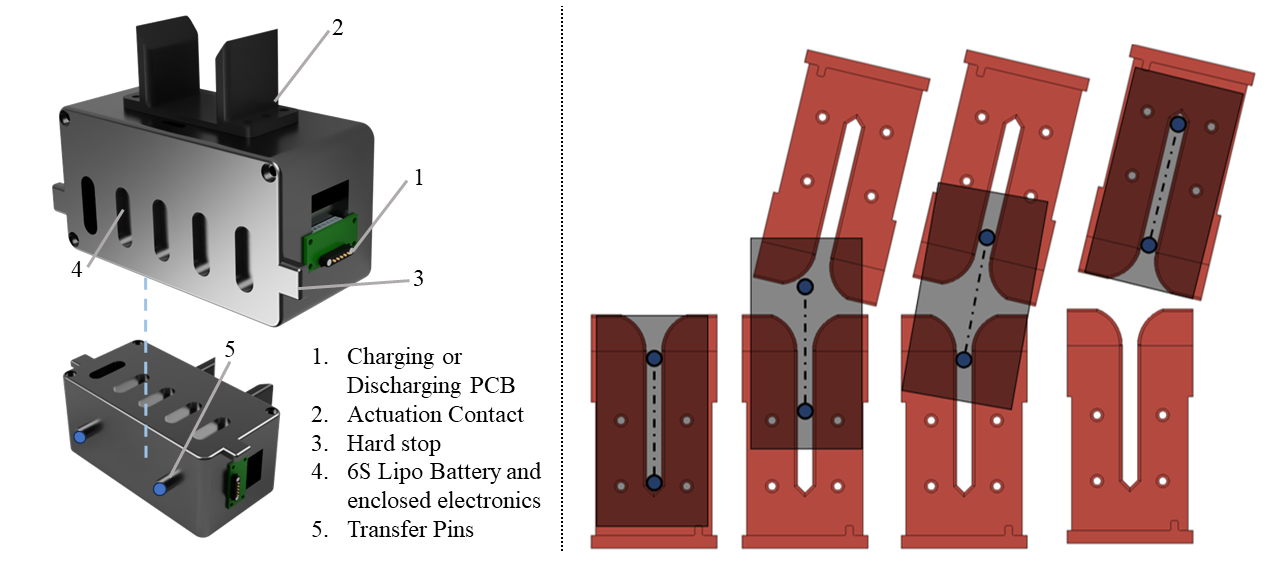}
    \caption{Battery module breakdown. Left: Annotated diagram of battery module. Right: Pin-slot transfer guidance of a battery module (shaded black) between two terminals.}
    \label{fig:pin-slot}
    \vspace{-1.0em}
\end{figure*}

Each battery terminal on the rover and hub features a guide slot with a fanned opening. Battery modules engage with these slots via two guide pins protruding from their bottom face. Fig.~\ref{fig:pin-slot} illustrates how this pin-slot design passively accounts for lateral and angular misalignment between the hub and rover terminals using the principle of exact constraints. In this implementation, the geometry accommodates up to $\pm15\%$ of the module's body width in misalignment. 

The battery terminals are designed to form a tileable array, allowing simple expansion without additional actuation. In this implementation, 3 terminals are used to support 2 rovers; this can be generalized to $n_r+1$ terminals for $n_r$ rovers, as a free slot is needed to perform tray operations.


The battery-swapping sequence is informed by a combination of sensors, custom PCBs, and software. Limit switches located along each axis of the swapping system are used to home each stepper motor. All electrical interfaces use standardized pairs of custom connector boards. The connector boards dedicate two pins for detecting continuity when a pair of PCBs is properly mated together, giving electrical verification for every mating-demating cycle for rover docking and battery module swapping. Following software verification of mating, power transfer is enabled between the pair of boards.The connector boards also contain small magnets that provide around 2N of magnetic force. These were implemented to provide passive retention that prevents the battery module from unplugging during nominal operation, but can be easily disengaged by the shifting mechanism for battery transfer. 


\section{Testing} 
Quantitative testing was first performed for battery module swapping and later for full system testing. Initial testing was done in a lab environment on flat ground, while later testing was conducted on outdoor field terrain.

\begin{figure}
\centering
\includegraphics[width=.8\linewidth]{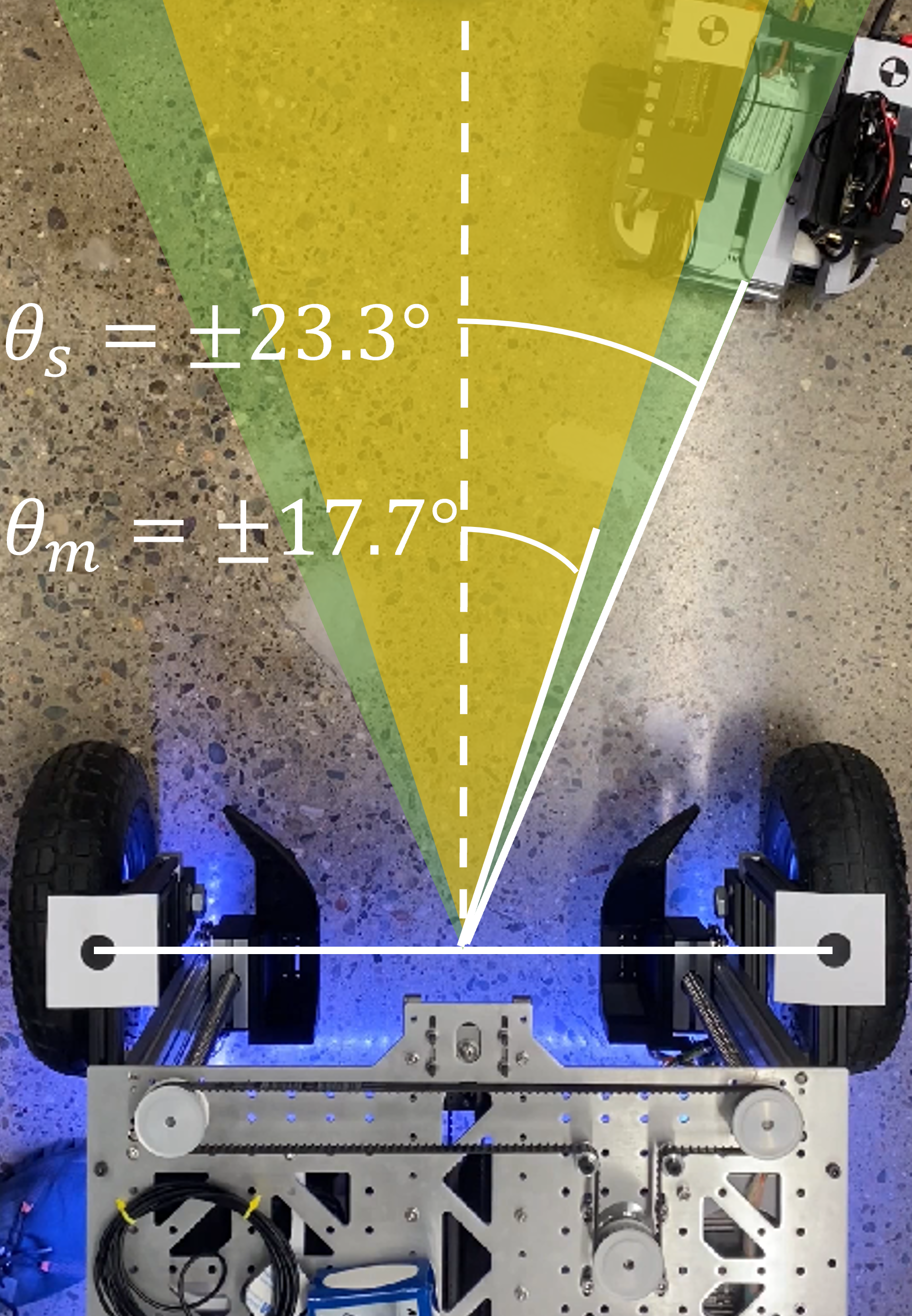}
    \caption{Evaluated yaw misalignment compensation for pure mechanical (yellow) and mechanical + software (green) control.}
    \label{yaw_misalign}
\end{figure}

First, the docking misalignment bounds evaluated through simulation were validated and quantified with repeated entry and lifting testing. The rover was placed at 25 different starting positions with respect to the hub robot and commanded to dock with and without software vision enabled to evaluate purely mechanical yaw compensation and the combined mechanical/software vision yaw compensation. From this testing, the mechanical yaw misalignment was characterized to be $\pm$17.72$\degree$ which increased to $\pm$23.72$\degree$ with software vision enabled as shown in Fig.~\ref{yaw_misalign}. 

The reliability of the battery swapping mechanism was evaluated by repeatedly exchanging the battery module between the rover and hub and cycling through each hub terminal. Electrical continuity was checked at each pass. The test was concluded at 50 successive battery swaps, yielding a 100\% success rate.

Next, full-stack integrated testing was performed within the indoor lab environment. In this testing, autonomous navigation, docking, swapping, and exiting were evaluated in sequence. Here, 15 consecutive dock-and-swaps were successfully executed without human interference, achieving an average servicing time of 98 seconds.

\begin{figure*} 
\centering
    \includegraphics[width=\linewidth]{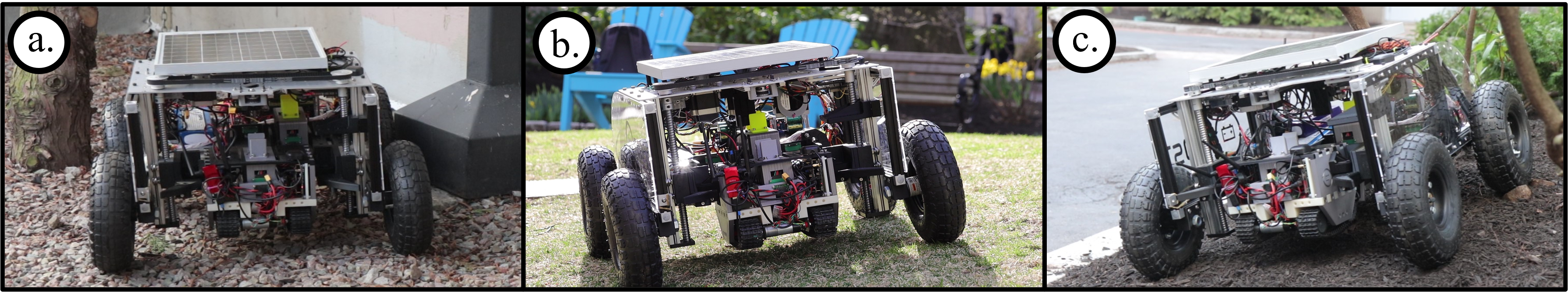}
    \caption{Successful docking carried out on varied surfaces: a) pebbles, b) grass, and c) mulch with $10\degree$ roll.}
    \label{IRL}
    \vspace{-1.0em}
\end{figure*}

Preliminary qualitative testing was conducted on outdoor terrain as shown in Fig.~\ref{IRL}. Docking was tested on grass, dirt, mulch, and pebbles; while not directly comparable to space-like regolith, these represented terrains of various textures, granularity, and friction coefficients. Due to time constraints, such field experiments were limited to a short testing window.

Outdoor testing was conducted on surfaces with a gradient of 10$\degree$ and 13$\degree$. Positioning the hub and rover in various orientations on the gradient enabled the success in pure pitch and roll conditions to be evaluated. Varying frontal pitch, docking was successful for the 10$\degree$ incline but not for the 13$\degree$ incline. For the large frontal pitch, the gravity vector was not sufficiently aligned with the lifting arms to properly engage the rover tabs. The battery swap procedure was unsuccessful for both the 10$\degree$ and 13$\degree$ roll tests. During the battery transfer step, the battery module's guide pins were observed slipping laterally at the gap between the rover and the hub. 

Geometric modifications may likely be sufficient to increase the tolerable operating range. By modifying the angle of mating surfaces, a larger range of pitch angles could be tolerated. By reducing the clearance between the battery terminals of the rover and hub, the pin slip failure mode can be eliminated. However, to broadly minimize risk, the roll and pitch angles of the hub should also be minimized prior to a docking attempt.


\section{Next Steps and Space Readiness}
With the aforementioned prototype system, an initial docking and swapping methodology was constructed and evaluated. However, this prototype was created as an earth-based system. Several further steps and considerations are required to adapt the system for space implementation, including environmental protection, hardware upgrades, and risk mitigation techniques.

\subsection{Environmental Considerations}
Depending on the planetary body, there are a variety of environmental considerations that need to be made in order to implement this architecture. These include dust, thermal, and radiation, which are challenges that persist across nearly all planetary bodies.

\subsubsection{Dust}
For planetary bodies such as Mars and the moon, dust poses a major concern for any exposed surfaces. To mitigate ingress concerns, all docking mechanisms shall be housed in hatches, which may be actuated open during the swapping procedure. To further protect the charging contacts, dust-resistant connectors such as fuzz buttons may be used in place of flat contact pad leads, enabling electrical contact despite dust buildup \cite{FuzzButtons}. Alternatively, wireless charging may be used to transfer power to and from the battery modules. Additionally, moving components shall be sealed with dust-resistant covers, such as bellows for lead screws and ceramics bearings for rotary joints, which have been successfully demonstrated on previous rovers \cite{dust}. 

\subsubsection{Thermal}
For planetary bodies with little to no atmosphere, heat fluxes incident on the rovers may swing drastically from day to night, making extreme temperatures a critical design consideration. Unlike Curiosity and Perseverance, the systems in this architecture do not benefit from the passive heat generation supplied by an RTG. Therefore, external heaters, akin to those on Spirit and Opportunity may be necessary. Furthermore, a cooling system may be necessary depending on the body's peak thermal radiation and its atmosphere's convection rate.

During a battery swap, it is critical for the rover to remain at working temperatures. Within the prototype system, a potential risk mitigation method was tested in which the rover maintains electrical operation when the battery is removed by switching to temporary support power. Through this, necessary heating or cooling systems may remain active during a swap. However, mission designs may elect to forgo this capability to reduce complexity. In this event, the rover will lose power during the battery swap and thereby cool down over time. 

Considering a battery swap occurring in a low-temperature environment (with no solar irradiation), the temperature of the rover as a function of time can be modeled using a simplified energy balance between the blackbody radiation emitted by the rover and thermal energy stored in the rover \cite{quattrocchi2022thermal}. The blackbody radiation emitted can be represented by: 

\begin{equation}
\frac{dE}{dt} = \epsilon\sigma A(T_{ambient}^4 - T_{rover}^4)
\label{EnergyBalance}
\end{equation}

where $\epsilon$ is the rover emissivity, $\sigma$ is the Stefan-Boltzmann constant, and A is the radiating surface area. Similarly, the change in thermal energy stored in a mass can be approximated by: 
\begin{equation}
\frac{dE}{dt} = mc\frac{dT}{dt}
\label{stored_energy}
\end{equation}
where $dT/dt$ is the instantaneous change in rover temperature, $m$ is the rover mass, and $c$ is the specific heat capacity of a material. 

From these equations, it can be shown that the amount of time for a Sojourner class (11 kg, 1.302 m$^2$, aluminum) rover to radiatively cool from the nominal operating temperature of $40\degree C$ to the lower electronics operating temperature of $-40\degree C$ \cite{Beauchamp} would take about 40 minutes; for rovers of larger size and thermal mass, this would take longer. 

In addition to the rover, maintaining a valid thermal environment for its battery module is critical for proper function. Equations \ref{EnergyBalance} and \ref{stored_energy} are again used to model the temperature of the prototype battery modules (0.4kg, 776cm$^2$, lithium ion, Cp = 960 J/kg$\cdot$K \cite{maleki1999thermal}). The battery module would take 8 minutes to cool from $40\degree C$ to 0$\degree C$, a conservative lower operating temperature limit for flight batteries. 

Based on our prototype earth-based implementation, a battery swap takes place in around one minute, providing a large margin before critical temperatures are reached for either a rover or individual battery module. In a space-optimized system, should radiative cooling during a battery swap be a concern, a thermal shroud may be implemented to enclose the region where the swap takes place. Additionally, if the heat dissipation rate of the battery module becomes a design constraint, additional auxiliary heaters could be enclosed within the battery module to maintain satisfactory thermal conditions.

\subsubsection{Ionizing Radiation}
As is standard for existing missions, electrical components should be shielded or hardened to protect against solar radiation. Regarding the design of space-ready modules, prolonged exposure to radiation can lead to the degradation of the performance of lithium ion batteries. This can be avoided by building the battery with its electrolyte, cathode active material, and binder optimized for radiation tolerance. NCM811 cathode material, PVDF as the binder, and LiBf4-based electrolyte are the materials that result in the least chemical degradation of the cells \cite{BattRad}. Metal oxide and nanocomposite coatings have also been shown to improve radiation resistance when applied to the surfaces of the cell, electrode, or other rover materials \cite{BattRad}. 

\subsection{Battery Module Design}

Efficient and energy-dense rechargeable batteries are crucial for increasing the range of planetary surface exploration. The batteries must have low operating temperatures, high reliability, and high energy density while minimizing mass and volume. They should also have low self-discharge and high coulombic efficiency and maintain performance after many charging cycles.
\begin{table}
\setlength{\tabcolsep}{10pt}
\begin{tabular}{|p{0.35\linewidth} | p{0.5\linewidth}|}
\hline
\textbf{Risk} & \textbf{Mitigation}\\
\hline\hline
Large misalignment between hub and rover before docking. & Physical bumpers guide rover into position. "Lift" phase of docking ensures gravitational alignment of rovers. If mobile hub is utilized, hub can relocate to location within gradient limits. \\
\hline
Battery module jams during swap. & Pin-track guides reduce the possibility of jams. Overcurrent detection on motors reverses and reattempts. \\
\hline
Loss of data during swap. & Auxiliary power or supercapacitors ensure the rover does not shut down when the battery is removed. \\
\hline
Failure of auxiliary power during swap. & Rover thermals will stay warm for up to 40 minutes. All swapping motors are relegated to the hub. Automatic rover reboot sequence once the new battery is serviced. \\
\hline
Dust impedes mechanical or electrical swap operations. & Fuzz-buttons or wireless charging lower risk of dust buildup. Hatch seals battery module port and docking bay between swaps. Bellows cover swapping mechanisms.  \\
\hline
Total failure of docking port. & One hub may have two or more redundant docking ports.  \\
\hline
Total failure of power generator. & Multiple hubs may be included in the network with a universal swapping architecture. \\
\hline
Rover is unable to return to the hub before losing power. & Astronauts or servicing robots may carry charged modules to spent rovers in the field\\
\hline
\end{tabular}
\caption{Flight Risk Mitigation Techniques}
\label{riskmitigation}
\vspace{-1.5em}
\end{table}

For rovers on the surface of Mars, ideal rechargeable battery should have a temperature operating range of -40$\degree C$ to 40$\degree C$, a capacity of at least 5-10Ah, a discharge rate of C/5-C/2, a specific energy greater than 100 Wh/kg, and an energy density of 120-160Wh/I \cite{BatteryReview}. Common chemical compositions of batteries include NiH2, NiCd, and AgZn, but lithium ion batteries are most commonly used as they can better meet the specifications previously discussed while minimizing mass, volume, and cost \cite{BatteryReview}. Higher-density batteries may be available in the future, such as lithium sulfur (Li-S), batteries, which have up to 5 times the energy density \cite{Li-S}. Solid-state batteries are also being developed for space applications by focusing on their performance at low temperatures.

\subsection{Risk Mitigation Strategies}
The proposed battery module system incurs several risks due to the nature of added hardware requirements and management logistics. Several risks and proposed mitigations are offered in table \ref{riskmitigation}.

\section{Conclusion}
Within this work, the possibility, benefits, limitations, and implementation of a battery-swapping system for supporting large fleets of rovers are evaluated. Given recent NASA milestones in lunar fission reactors, a new paradigm of outsourced power generation may enable the streamlining of rovers. While this may enable dozens of lower-cost rovers to operate off a shared power source, there are a variety of logistic limitations, such as a maximum number of serviced rovers, and a limited area of coverage. 

In this work, a prototype system was implemented to demonstrate several of the core technologies required for the docking and battery-swapping process. From this, a functional methodology was constructed and validated with earth-based testing. In future research, heavy focus on the dust-proofing of the system should be implemented due to the high number of interacting parts and electrical contacts. Further trade studies evaluating the cost-benefit analysis of battery swapping versus wired or wireless charging should be conducted. We hope that this yields future discussion as to the possibilities of high-wattage power infrastructure and multi-robot systems for the future of rover design.

\newpage{}

\appendices{}

\section{Physical Parameters of Prototype}
\label{sec:physparams}
For comparison with future systems, the physical parameters of the prototype system have been outlined in Table \ref{tab:boostspecs}

\renewcommand{\arraystretch}{1.1}
\begin{table} [h!]
    \centering
    \begin{tabular}{|p{4cm}|p{2.5cm}|p{0.75cm}|}
    \hline 
    \textbf{Parameter} & \textbf{Value} & \textbf{Units}\\
    \hline 
    \hline 
         Rover Mass&  2.5& kg\\ \hline 
         Battery Module Mass&  0.4& kg\\ \hline 
         Hub Mass&  15& kg\\ \hline 
         Rover Payload Capacity& 2.5 &kg \\ \hline
         
         Rover Dimensions & 260 x 220 x 331 & mm \\ \hline 
         Hub Dimensions & 655 x 340 x 660 & mm \\ \hline 
         Battery Module Dimensions & 160 x 125 x 66 & mm\\ \hline 
         
         Rover Velocity & 1 & m/s\\ \hline 
         Hub Velocity & 2.5 & m/s \\ \hline 
         
        Yaw Allowance with only Mechanical Compensation  & 17.72 & \degree \\ \hline
         Yaw Allowance with Autonomous Navigation and Mechanical Compensation & 23.72 & \degree \\ \hline 
         
         Hub Lifting Torque & 0.0179 & N$\cdot$m \\ \hline
         Hub Shifter Torque & 0.445 & N$\cdot$m \\ \hline
         Hub Indexer Torque & 0.01226 & N$\cdot$m \\ \hline 
         Force required to overcome magnet and swap battery  & 4.453 & N \\ \hline
         Battery Module Capacity & 2.8 & AHr \\ \hline
         Battery Module Chargetime & 1 & hr \\ \hline
         Average Rover Runtime & 2 & hr \\ \hline
         Battery Swap Success Rate (50 consecutive tests) & 100 & \% \\ \hline
         Battery Module Max Hold/Slip Angle & 25 & \degree \\ \hline        
         
    \end{tabular}
    \caption{Key design specifications and performance parameters for prototype system}
    \label{tab:boostspecs}
\end{table}
\newpage
\section{Labeled Diagrams of Prototype}
\begin{figure}[h!]
\centering
    \includegraphics[width=0.85\linewidth]{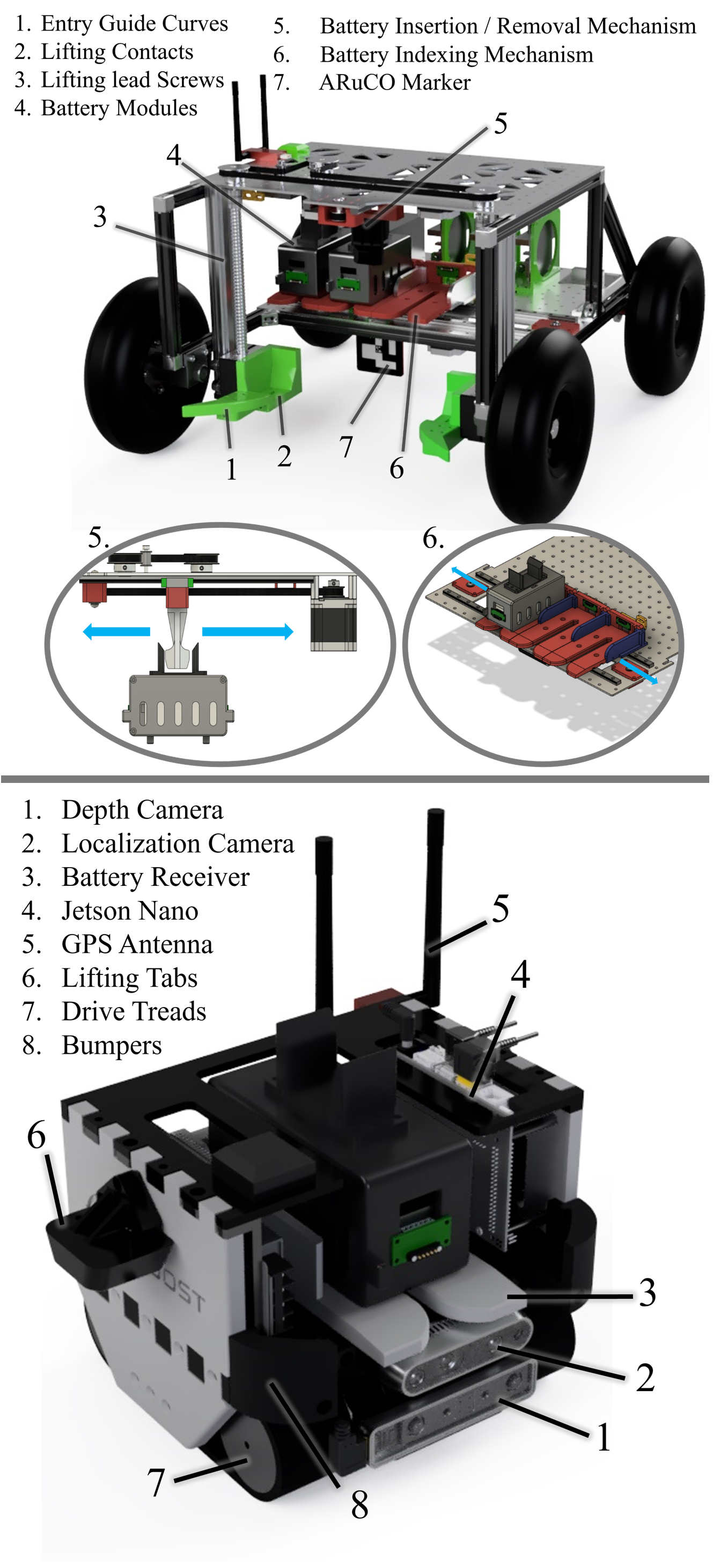}
    \caption{Diagram of hub (top) and rover (bottom) with major subsystems and components labeled.}
    \label{RoverPic}
\end{figure}

CAD renderings of the multi-agent battery swapping system prototype are shown in Figure \ref{RoverPic}.
\label{sec:Renders}

\newpage{}
\section{Hub Network Coverage}
\label{sec:hubcoverage}
Centralized power systems impose travel constraints on the network, detailed as follows.
\begin{figure} [h]
    \includegraphics[width=\linewidth]{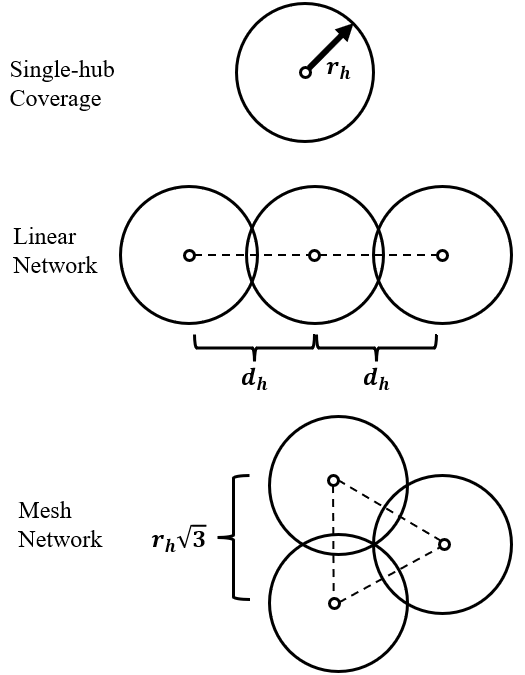}
    \caption{Servicing area for single or many hub robots}
    \label{Hub Coverage}
\end{figure}

With an area of coverage mapped, rover paths can be planned within the network, subject to the following two constraints:
\begin{enumerate}
    \item Each rover trajectory must start and end at a hub
    \item The maximum length of a rover trajectory is $2R_h$
\end{enumerate}

\subsection{Single-hub coverage}
The maximum distance a rover can travel away from a hub is characterized by its battery life its power consumption. The battery life per rover, $t_m$, can be represented as
$$t_m=\frac{Q_bV_b}{P_m}$$
where $Q_b$ is the battery capacity, $V_b$ is the voltage of a battery module, and $P_m$ is the average power consumption per rover. Considering a rover driving out from the hub until half capacity is reached and returning with the remaining energy, a maximum travel distance, $R_h$, can be determined:
$$R_h=\frac{1}{2}v_mt_m$$
Where $v_m$ is the velocity of a rover. Sweeping this travel distance across all starting angles, a hub's servicing area is created. This can be expressed by 

$$A_{h}=\pi R_h^2$$
and can be visualized in Fig.~\ref{Hub Coverage}. 

\subsection{Multi-hub coverage}
To increase the area of coverage, support a greater number of rovers, and add redundancy, additional hubs may be added to the network at a distance $d_h$ away from an existing hub. If hubs are placed at a distance of $d_h<2R_h$, an overlapping region is created, enabling intra-hub rover travel. The area of overlap for two hubs can thus be defined as 

$$A_{overlap}=2R_h^2\cos^{-1}\left({\frac{d_h}{2R_h}}\right)-\frac{1}{2}d_h\sqrt{4R_h^2-d_h^2}.$$

To maximize the length of a network, hubs can be positioned as links in a chain formation, in which they only overlap with one previous and following hub. In this case, the total area of the network, $A_{coverage}$, can be computed as 

$$A_{coverage}=n_h(A_h-A_{overlap})+A_{overlap}$$

where $n_h$ is the number of hubs in the chain network.

Alternatively, if servicing a wide region is desired, the hubs be tiled in a form resembling hexagonal packing. To eliminate gaps in coverage, a set of Johnson circles can be created, in which the Johnson triangle created by the three hubs is equilateral. Using the law of cosines, the maximum hub spacing for gapless coverage is determined as $d_h=r_h\sqrt{3}$. In this case, the total area of the network, $A_{coverage}$, can be computed as 

$$A_{coverage}=n_hA_h-\max(0, 2n_h-3)A_{overlap}$$

\section*{Acknowledgments}
The prototype in this study was designed, constructed and tested via a collaboration between Mechanical Engineering and Electrical Engineering Capstone student teams at Northeastern University.  The authors recognize Professors Michael Allshouse, Andrew Gouldstone, and Bahram Shafai for their coordination of these courses. The authors would additionally like to thank Professor Mehdi Abedi for advising the mechanical development of the prototype.

The authors would like to thank Northeastern University's MIE and EECE departments for providing funding to complete this work.

We would additionally like to thank Patrick DeGrosse Jr., from NASA's Jet Propulsion Laboratory, for providing technical feedback throughout the development of the project.

\newpage{}
\bibliographystyle{IEEEtran}
\bibliography{references}

\begin{thebibliography}{10}
\providecommand{\url}[1]{#1}
\csname url@samestyle\endcsname
\providecommand{\newblock}{\relax}
\providecommand{\bibinfo}[2]{#2}
\providecommand{\BIBentrySTDinterwordspacing}{\spaceskip=0pt\relax}
\providecommand{\BIBentryALTinterwordstretchfactor}{4}
\providecommand{\BIBentryALTinterwordspacing}{\spaceskip=\fontdimen2\font plus
\BIBentryALTinterwordstretchfactor\fontdimen3\font minus \fontdimen4\font\relax}
\providecommand{\BIBforeignlanguage}[2]{{%
\expandafter\ifx\csname l@#1\endcsname\relax
\typeout{** WARNING: IEEEtran.bst: No hyphenation pattern has been}%
\typeout{** loaded for the language `#1'. Using the pattern for}%
\typeout{** the default language instead.}%
\else
\language=\csname l@#1\endcsname
\fi
#2}}
\providecommand{\BIBdecl}{\relax}
\BIBdecl

\bibitem{artemis}
M.~Smith, D.~Craig, N.~Herrmann, E.~Mahoney, J.~Krezel, N.~McIntyre, and K.~Goodliff, ``The artemis program: An overview of nasa's activities to return humans to the moon,'' in \emph{2020 IEEE Aerospace Conference}, 2020, pp. 1--10.

\bibitem{artemisplan}
\BIBentryALTinterwordspacing
{National Aeronautics and Space Administration}. (2020) {Artemis Plan: NASA's Lunar Exploration Program Overview}. Accessed on October 6th, 2023. [Online]. Available: \url{https://www.nasa.gov/wp-content/uploads/2020/12/artemis_plan-20200921.pdf}
\BIBentrySTDinterwordspacing

\bibitem{Solar_rovers}
G.~Landis, ``Exploring mars with solar-powered rovers,'' in \emph{Conference Record of the Thirty-first IEEE Photovoltaic Specialists Conference, 2005.}, 2005, pp. 858--861.

\bibitem{curiosity}
R.~Welch, D.~Limonadi, and R.~Manning, ``Systems engineering the curiosity rover: A retrospective,'' in \emph{2013 8th International Conference on System of Systems Engineering}, 2013, pp. 70--75.

\bibitem{perserverence}
R.~Rieber, M.~McHenry, P.~Twu, and M.~M. Stragier, ``Planning for a martian road trip - the mars2020 mobility systems design,'' in \emph{2022 IEEE Aerospace Conference (AERO)}, 2022, pp. 01--18.

\bibitem{rtgefficiency}
\BIBentryALTinterwordspacing
R.~O’Brien, R.~Ambrosi, N.~Bannister, S.~Howe, and H.~Atkinson, ``Safe radioisotope thermoelectric generators and heat sources for space applications,'' \emph{Journal of Nuclear Materials}, vol. 377, no.~3, pp. 506--521, 2008. [Online]. Available: \url{https://www.sciencedirect.com/science/article/pii/S0022311508002420}
\BIBentrySTDinterwordspacing

\bibitem{tether}
A.~Barchowsky, A.~Amirahmadi, K.~Botteon, A.~Goddu, P.~M. Gregory~Carr, Curtis~Jin, S.~Sposato, and S.~Yang, ``Tethered power systems for lunar mobility and power transmission,'' in \emph{Lunar Surface Innovation Consortium, 1380-SP Monthly Meeting}, April 2022.

\bibitem{wireless1}
\BIBentryALTinterwordspacing
{Astrobotic Technology, Inc.} (2020) {High Power Near-Field Wireless Transfer for Dust Intensive Applications}. Accessed on October 4th, 2023. [Online]. Available: \url{https://sbir.nasa.gov/SBIR/abstracts/20/sbir/phase2/SBIR-20-2-Z13.02-6350.html}
\BIBentrySTDinterwordspacing

\bibitem{wireless2}
\BIBentryALTinterwordspacing
{Yank Technologies, Inc.} (2023) {High Power Wireless Charging Systems for Extreme Environments }. Accessed on October 4th, 2023. [Online]. Available: \url{https://sbir.nasa.gov/SBIR/abstracts/23/sbir/phase1/SBIR-23-1-Z13.05-1830.html}
\BIBentrySTDinterwordspacing

\bibitem{kilopower}
M.~Gibson, D.~Poston, P.~Mcclure, T.~Godfroy, J.~Sanzi, and M.~Briggs, ``The kilopower reactor using stirling technology (krusty) nuclear ground test results and lessons learned,'' 07 2018.

\bibitem{FissionAwards}
T.~A. Tofil, ``Overview of nasa's fission surface power (fsp) project,'' in \emph{Nuclear Technical Discipline Meeting}, 2022.

\bibitem{rtg}
J.~Kempenaar, K.~Novak, M.~Redmond, E.~Farias, K.~Singh, and M.~Wagner, ``Detailed surface thermal design of the mars 2020 rover,'' in \emph{48th International Conference on Environmental Systems}, 2018.

\bibitem{SpiritandOpportunityPower}
\BIBentryALTinterwordspacing
B.~V. Ratnakumar, M.~C. Smart, R.~C. Ewell, L.~D. Whitcanack, K.~B. Chin, and S.~Surampudi, ``{Lithium-Ion rechargeable batteries on Mars Rover},'' 2004. [Online]. Available: \url{https://hdl.handle.net/2014/38818}
\BIBentrySTDinterwordspacing

\bibitem{CADRE}
\BIBentryALTinterwordspacing
F.~Rossi and J.-P. de~la Croix, ``Cadre: A lunar technology demo of multi-agent autonomy enabling distributed measurements,'' \emph{IEEE SMC}, 2023. [Online]. Available: \url{https://ieeesmc2023.org/abstract_files/SMC23_1581_FI.pdf}
\BIBentrySTDinterwordspacing

\bibitem{isru}
\BIBentryALTinterwordspacing
G.~Sanders, ``In situ resource utilization (isru) - surface excavation \& construction,'' \emph{NASA Advisory Council Technology, Innovation and Engineering Committee}, 2021. [Online]. Available: \url{https://www.nasa.gov/wp-content/uploads/2015/03/jsanders_lunar_isru_tagged_0.pdf}
\BIBentrySTDinterwordspacing

\bibitem{astroboticwireless}
\BIBentryALTinterwordspacing
``Astrobotic’s wireless charging system for the moon can survive lunar night 2022,'' Jun 2022. [Online]. Available: \url{https://www.astrobotic.com/astrobotics-wireless-charging-system-for-the-moon-can-survive-lunar-night/}
\BIBentrySTDinterwordspacing

\bibitem{ChargeProtocol}
\BIBentryALTinterwordspacing
S.~S. Zhang, ``The effect of the charging protocol on the cycle life of a li-ion battery,'' \emph{Journal of Power Sources}, vol. 161, no.~2, pp. 1385--1391, 2006. [Online]. Available: \url{https://www.sciencedirect.com/science/article/pii/S0378775306011839}
\BIBentrySTDinterwordspacing

\bibitem{ChargeRate}
\BIBentryALTinterwordspacing
C.~Zhang, J.~Jiang, Y.~Gao, W.~Zhang, Q.~Liu, and X.~Hu, ``Charging optimization in lithium-ion batteries based on temperature rise and charge time,'' \emph{Applied Energy}, vol. 194, pp. 569--577, 2017. [Online]. Available: \url{https://www.sciencedirect.com/science/article/pii/S0306261916315033}
\BIBentrySTDinterwordspacing

\bibitem{EVFastCharge}
\BIBentryALTinterwordspacing
W.~Xie, X.~Liu, R.~He, Y.~Li, X.~Gao, X.~Li, Z.~Peng, S.~Feng, X.~Feng, and S.~Yang, ``Challenges and opportunities toward fast-charging of lithium-ion batteries,'' \emph{Journal of Energy Storage}, vol.~32, p. 101837, 2020. [Online]. Available: \url{https://www.sciencedirect.com/science/article/pii/S2352152X20316741}
\BIBentrySTDinterwordspacing

\bibitem{batterymanagement1}
N.~Kamra, T.~K.~S. Kumar, and N.~Ayanian, ``Combinatorial problems in multirobot battery exchange systems,'' \emph{IEEE Transactions on Automation Science and Engineering}, vol.~15, no.~2, pp. 852--862, 2018.

\bibitem{batterymanagement2}
X.~Liu, T.~Zhao, S.~Yao, C.~B. Soh, and P.~Wang, ``Distributed operation management of battery swapping-charging systems,'' \emph{IEEE Transactions on Smart Grid}, vol.~10, no.~5, pp. 5320--5333, 2019.

\bibitem{batteryswap1}
\BIBentryALTinterwordspacing
J.~Wu, G.~Qiao, J.~Ge, H.~Sun, and G.~Song, ``Automatic battery swap system for home robots,'' \emph{International Journal of Advanced Robotic Systems}, vol.~9, no.~6, p. 255, 2012. [Online]. Available: \url{https://doi.org/10.5772/54025}
\BIBentrySTDinterwordspacing

\bibitem{batteryswap2}
J.~Zhang, G.~Song, Y.~Li, G.~Qiao, and Z.~Li, ``Battery swapping and wireless charging for a home robot system with remote human assistance,'' \emph{IEEE Transactions on Consumer Electronics}, vol.~59, no.~4, pp. 747--755, 2013.

\bibitem{batteryswap3}
Y.~Saito, K.~Asai, C.~Yongwoon, T.~Iyota, K.~Watanabe, and Y.~Kubota, ``Development of a battery support system for the prolonged activity of mobile robots,'' \emph{Ieej Transactions on Electronics, Information and Systems}, vol. 128, pp. 1557--1566, 10 2008.

\bibitem{batteryswap4}
Y.-C. Wu, M.-C. Teng, and Y.-J. Tsai, ``Robot docking station for automatic battery exchanging and charging,'' in \emph{2008 IEEE International Conference on Robotics and Biomimetics}, 2009, pp. 1043--1046.

\bibitem{uavswap}
K.~A. Swieringa, C.~B. Hanson, J.~R. Richardson, J.~D. White, Z.~Hasan, E.~Qian, and A.~Girard, ``Autonomous battery swapping system for small-scale helicopters,'' in \emph{2010 IEEE International Conference on Robotics and Automation}, 2010, pp. 3335--3340.

\bibitem{uavswap2}
N.~K. Ure, G.~Chowdhary, T.~Toksoz, J.~P. How, M.~A. Vavrina, and J.~Vian, ``An automated battery management system to enable persistent missions with multiple aerial vehicles,'' \emph{IEEE/ASME Transactions on Mechatronics}, vol.~20, no.~1, pp. 275--286, 2015.

\bibitem{rimfax}
S.-E. Hamran, D.~Paige, H.~E.~F. Amundsen, T.~Berger, S.~Brovoll, L.~Carter, L.~Damsgard, H.~Dypvik, J.~Eide, S.~Eide, R.~Ghent, N.~Helleren, J.~Kohler, M.~Mellon, D.~Nunes, D.~Plettemeier, K.~Rowe, P.~Russell, and M.~N{\O}yan, ``Radar imager for mars’ subsurface experiment—rimfax,'' \emph{Space Science Reviews}, vol. 216, 12 2020.

\bibitem{instruments}
A.~R. Vasavada, ``Mission overview and scientific contributions from the mars science laboratory curiosity rover after eight years of surface operations.'' \emph{Space Science Reviews}, vol.~14, p. 218, 04 2022.

\bibitem{ROS2}
\BIBentryALTinterwordspacing
S.~Macenski, T.~Foote, B.~Gerkey, C.~Lalancette, and W.~Woodall, ``Robot operating system 2: Design, architecture, and uses in the wild,'' vol.~7, no.~66, 2022, p. eabm6074. [Online]. Available: \url{https://www.science.org/doi/abs/10.1126/scirobotics.abm6074}
\BIBentrySTDinterwordspacing

\bibitem{NAV2}
S.~Macenski, T.~Moore, D.~Lu, A.~Merzlyakov, and M.~Ferguson, ``From the desks of ros maintainers: A survey of modern \& capable mobile robotics algorithms in the robot operating system 2,'' \emph{Robotics and Autonomous Systems}, 2023.

\bibitem{garrido2014automatic}
S.~Garrido-Jurado, R.~Mu{\~n}oz-Salinas, F.~J. Madrid-Cuevas, and M.~J. Mar{\'\i}n-Jim{\'e}nez, ``Automatic generation and detection of highly reliable fiducial markers under occlusion,'' \emph{Pattern Recognition}, vol.~47, no.~6, pp. 2280--2292, 2014.

\bibitem{OpenCV}
G.~Bradski, ``{The OpenCV Library},'' \emph{Dr. Dobb's Journal of Software Tools}, 2000.

\bibitem{EKF}
F.~E. Daum, ``Extended kalman filters,'' in \emph{Springer Reference}.\hskip 1em plus 0.5em minus 0.4em\relax Springer, 2015.

\bibitem{baydas2019defining}
S.~Baydas and B.~Karakas, ``Defining a curve as a bezier curve,'' \emph{Journal of Taibah University for Science}, vol.~13, no.~1, pp. 522--528, 2019.

\bibitem{convhull}
\BIBentryALTinterwordspacing
C.~B. Barber, D.~P. Dobkin, and H.~Huhdanpaa, ``The quickhull algorithm for convex hulls,'' \emph{ACM Trans. Math. Softw.}, vol.~22, no.~4, p. 469–483, dec 1996. [Online]. Available: \url{https://doi.org/10.1145/235815.235821}
\BIBentrySTDinterwordspacing

\bibitem{FuzzButtons}
\BIBentryALTinterwordspacing
I.~Townsend, R.~Mueller, and A.~Dokos, ``Self-cleaning filament connector,'' 04 2017, u.S. Patent 9620888B2. [Online]. Available: \url{https://patents.google.com/patent/US9620888B2/}
\BIBentrySTDinterwordspacing

\bibitem{dust}
\BIBentryALTinterwordspacing
K.~M. Cannon, C.~B. Dreyer, G.~F. Sowers, J.~Schmit, T.~Nguyen, K.~Sanny, and J.~Schertz, ``Working with lunar surface materials: Review and analysis of dust mitigation and regolith conveyance technologies,'' \emph{Acta Astronautica}, vol. 196, pp. 259--274, 2022. [Online]. Available: \url{https://www.sciencedirect.com/science/article/pii/S0094576522001965}
\BIBentrySTDinterwordspacing

\bibitem{quattrocchi2022thermal}
G.~Quattrocchi, A.~Pittari, M.~D. Dalla~Vedova, and P.~Maggiore, ``The thermal control system of nasa’s curiosity rover: a case study,'' in \emph{IOP Conference Series: Materials Science and Engineering}, vol. 1226, no.~1.\hskip 1em plus 0.5em minus 0.4em\relax IOP Publishing, 2022, p. 012113.

\bibitem{Beauchamp}
R.~Surampudi, J.~Blosiu, P.~Stella, J.~Elliott, J.~Castillo, T.~Yi, J.~Lyons, M.~Piszczor, J.~McNatt, C.~Taylor \emph{et~al.}, ``Solar power technologies for future planetary science missions,'' \emph{Jet Propulsion Lab, California Institute of Technology}, 2017.

\bibitem{maleki1999thermal}
H.~Maleki, S.~Al~Hallaj, J.~R. Selman, R.~B. Dinwiddie, and H.~Wang, ``Thermal properties of lithium-ion battery and components,'' \emph{Journal of the Electrochemical Society}, vol. 146, no.~3, p. 947, 1999.

\bibitem{BattRad}
\BIBentryALTinterwordspacing
Y.~Gao, F.~Qiao, W.~Hou, L.~Ma, N.~Li, C.~Shen, T.~Jin, and K.~Xie, ``Radiation effects on lithium metal batteries,'' \emph{The Innovation}, vol.~4, no.~4, p. 100468, 2023. [Online]. Available: \url{https://www.sciencedirect.com/science/article/pii/S2666675823000966}
\BIBentrySTDinterwordspacing

\bibitem{BatteryReview}
\BIBentryALTinterwordspacing
A.~D. Pathak, S.~Saha, V.~K. Bharti, M.~M. Gaikwad, and C.~S. Sharma, ``A review on battery technology for space application,'' \emph{Journal of Energy Storage}, vol.~61, p. 106792, 2023. [Online]. Available: \url{https://www.sciencedirect.com/science/article/pii/S2352152X23001895}
\BIBentrySTDinterwordspacing

\bibitem{Li-S}
B.~Samaniego, E.~Carla, L.~O’Neill, and M.~Nestoridi, ``High specific energy lithium sulfur cell for space application,'' \emph{E3S Web of Conferences}, vol.~16, p. 08006, 01 2017.

\end{thebibliography}
\section*{Biography}
\begin{biographywithpic}
{Ethan Holand}{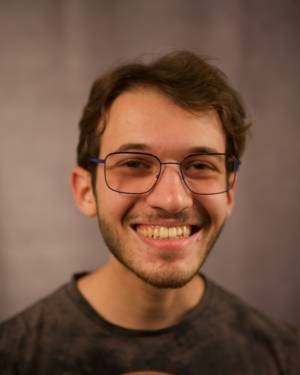}
is a graduate student at Carnegie Mellon University pursuing a Master of Science in Robotics. He previously completed his Bachelor's in Mechanical Engineering at Northeastern University. He has recently completed a co-op and internship at NASA's Jet Propulsion Laboratory within the Spacecraft Mechanical Engineering section, where he worked on impact testing for the landing gear of the Mars Sample Return mission. His research interests include locomotion and multi-agent systems for space robotics.
\end{biographywithpic}
\begin{biographywithpic}
{Jarrod Homer}{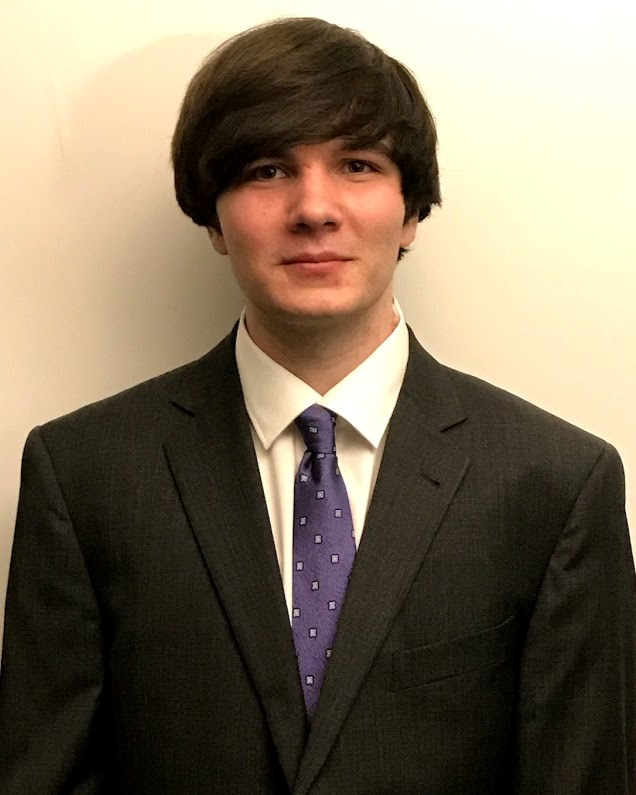}
is a current graduate student at Northeastern University, pursuing a Master's degree in Robotics with a concentration in Electrical and Computer Engineering. He holds a Bachelor's degree in Electrical Engineering from the same institution. He has 8 months of internship experience at Amazon Robotics working on the electrical design of sensors. His research interests include novel system engineering, electronics, and control. 
\end{biographywithpic}
\begin{biographywithpic}
{Alex Storrer}{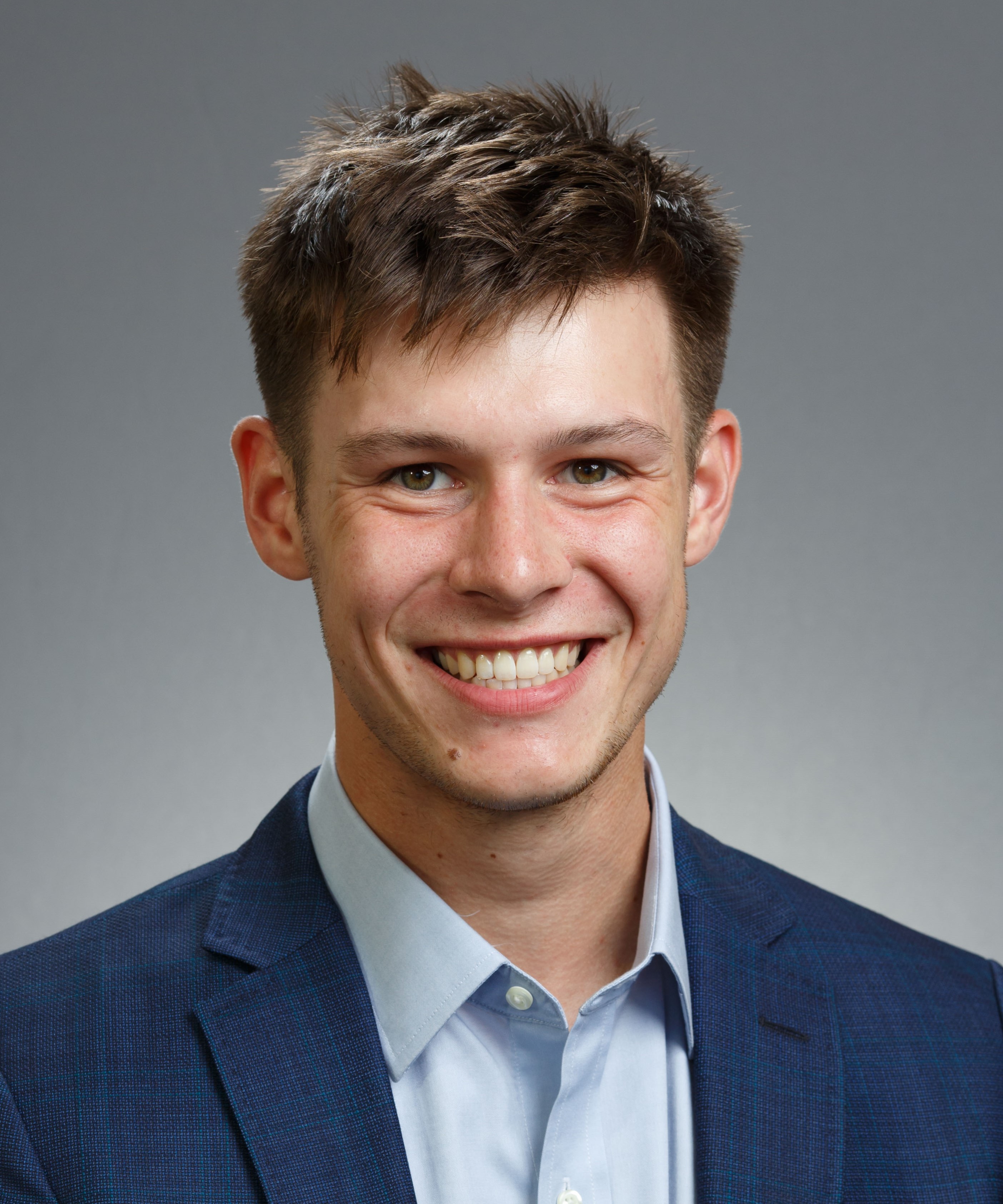}
is an undergraduate Mechanical Engineering \& Physics student at Northeastern University. He has completed two mechanical engineering internships at SpaceX within the propulsion department, and he currently is working as a mechanical engineer at MIT's Plasma Science and Fusion Center. He has multi-phase flow and high-speed aerodynamics research experience and plans to pursue a graduate degree studying turbulent fluid mechanics. He also enjoys robotics on the side. 
\end{biographywithpic}
\begin{biographywithpic}
{Musheera Khandeker}{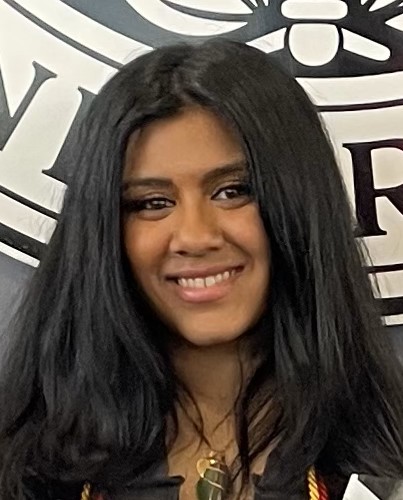}
completed her bachelor's degree in Electrical Engineering from Northeastern University, and is currently working as a Mixed Signal Design Engineer at Draper. Her past internship experience includes NK Labs, and Acacia Communications as an electrical engineer. Her research interests include soft robotics and EV low-voltage systems. 
\end{biographywithpic}
\begin{biographywithpic}
{Ethan F. Muhlon}{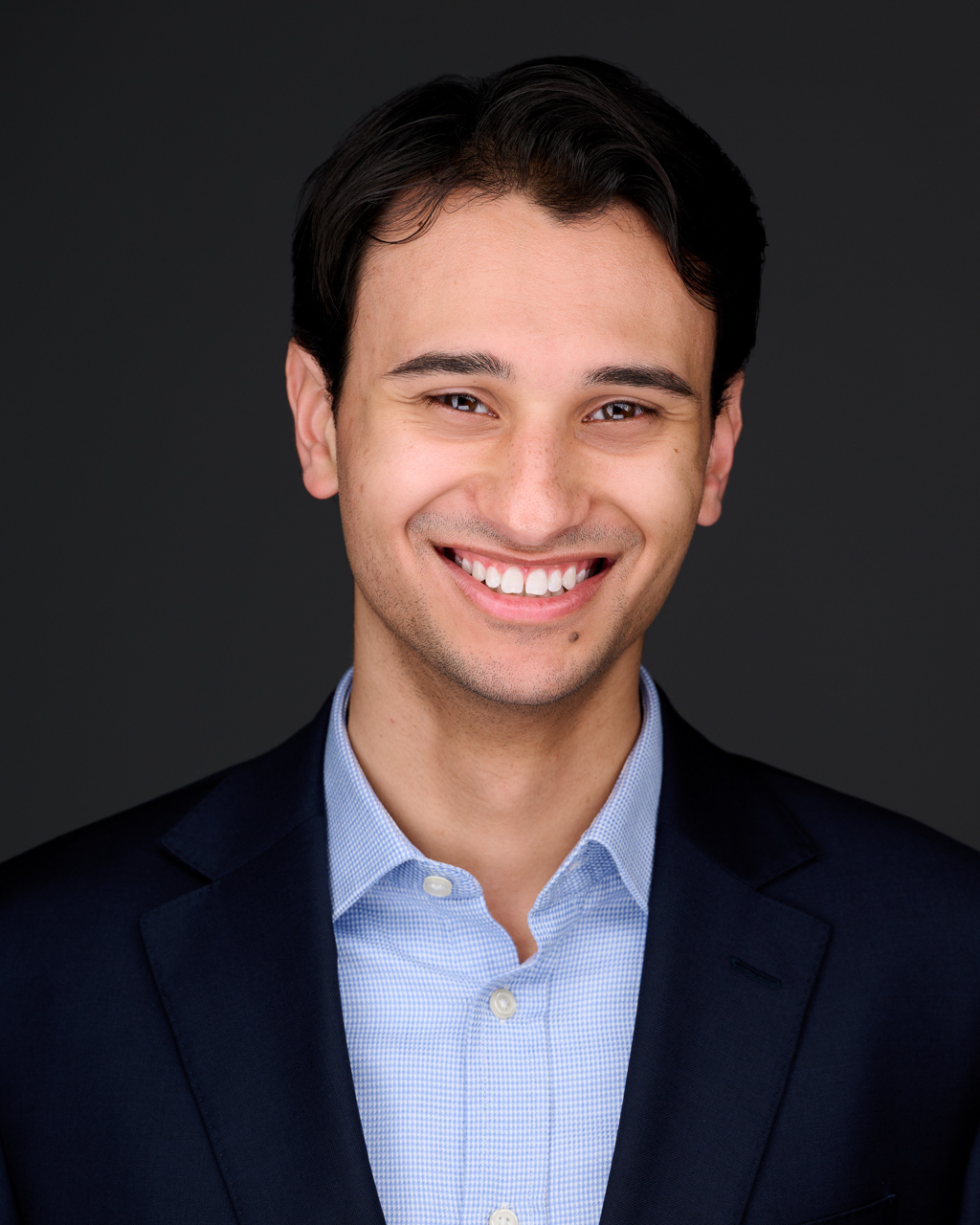}
received his Bachelor's degree in Mechanical Engineering from Northeastern University with a minor in Robotics. He currently works as a mechanical engineer at neuro42 developing portable MRI machines paired with a robotic assistant. Ethan has experience working at medical device companies and with the DoD. His research interests include medical and surgical robotics.
\end{biographywithpic}
\begin{biographywithpic}
{Maulik Patel}{bio/maulik.jpg}
received his Bachelor's degree in Electrical Engineering from Northeastern University, and is currently pursuing a Master of Science in Robotics. He has worked as an RF engineer co-op at CPI, and as a Electrical Research and Development engineer at Werfen. Beyond this, his engineering interests include EV racing and autonomous field robotics.   
\end{biographywithpic}
\begin{biographywithpic}
{Ben-oni Vainqueur}{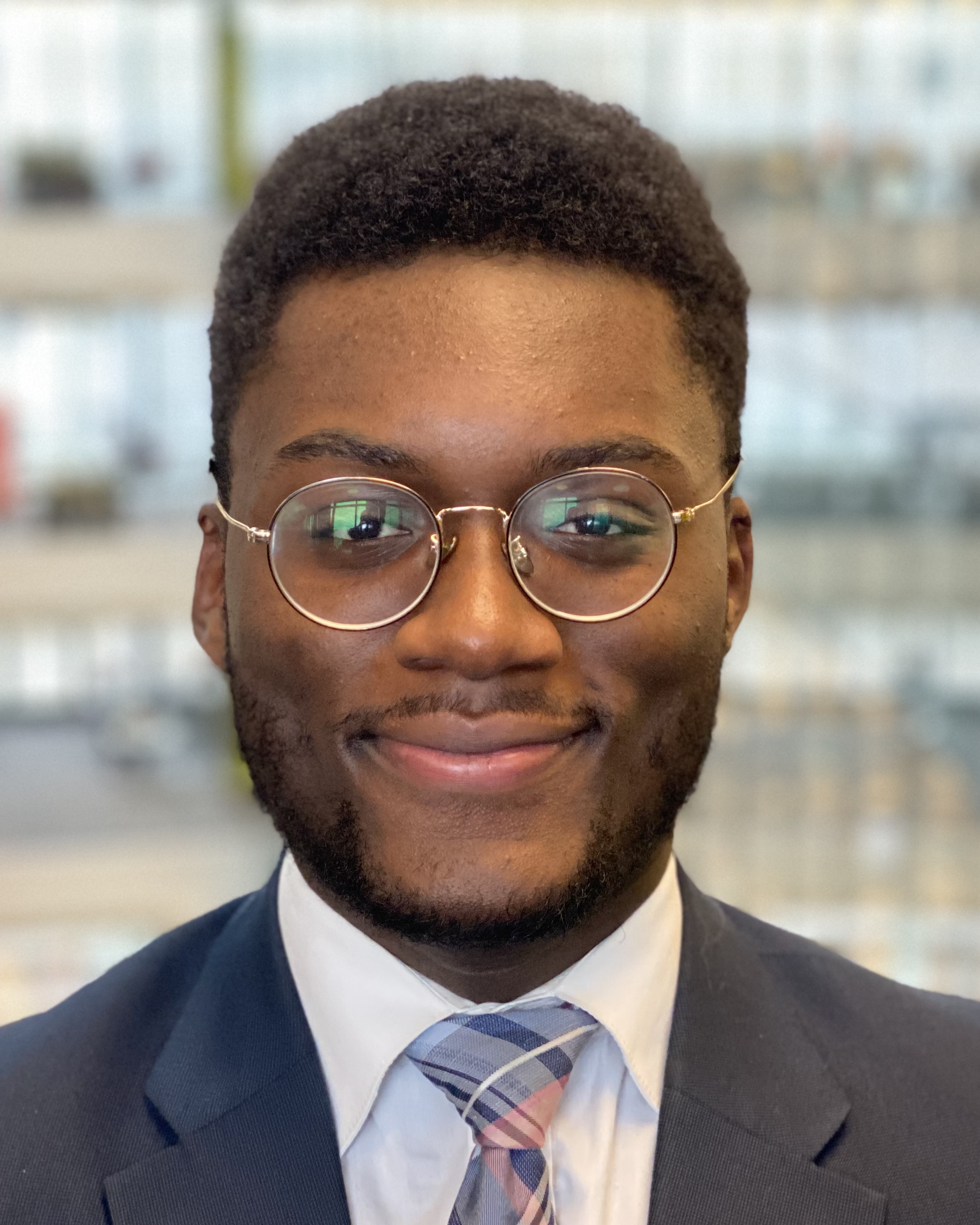}
received his Bachelor's degree in Computer Engineering and Computer Science from Northeastern University, and is pursuing a Master's Degree in Electrical and Computer Engineering, with a concentration in machine intelligence. He has worked as a software/automation engineer at Amazon Robotics, a software/firmware developer at Microsoft, and is currently a software engineer and researcher at the DOE Idaho National Laboratory. His engineering interests include robotics and machine learning. 
\end{biographywithpic}
\begin{biographywithpic}
{David Antaki}{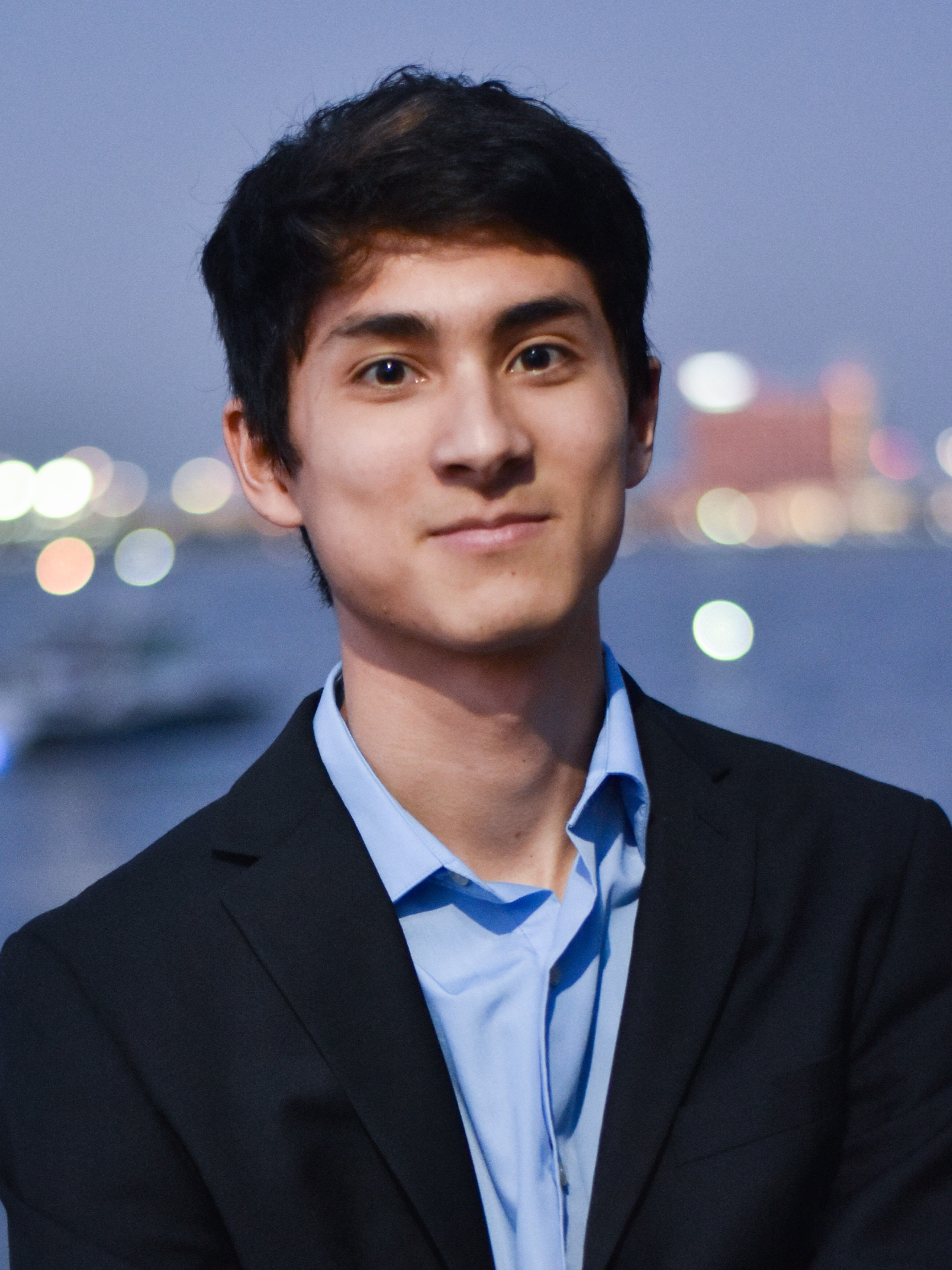}
received his Bachelor's degree in Computer Engineering and Computer Science from Northeastern University. He has worked at SpaceX as a Starship Engineer, a Firmware Engineer at Inkbit, and a Firmware Engineer at REEKON Tools. David is currently a software engineer at Applied Intuition. His engineering and research interests include applying deep learning to robotic systems for human-like behavior.
\end{biographywithpic}
\begin{biographywithpic}
{Naomi Cooke}{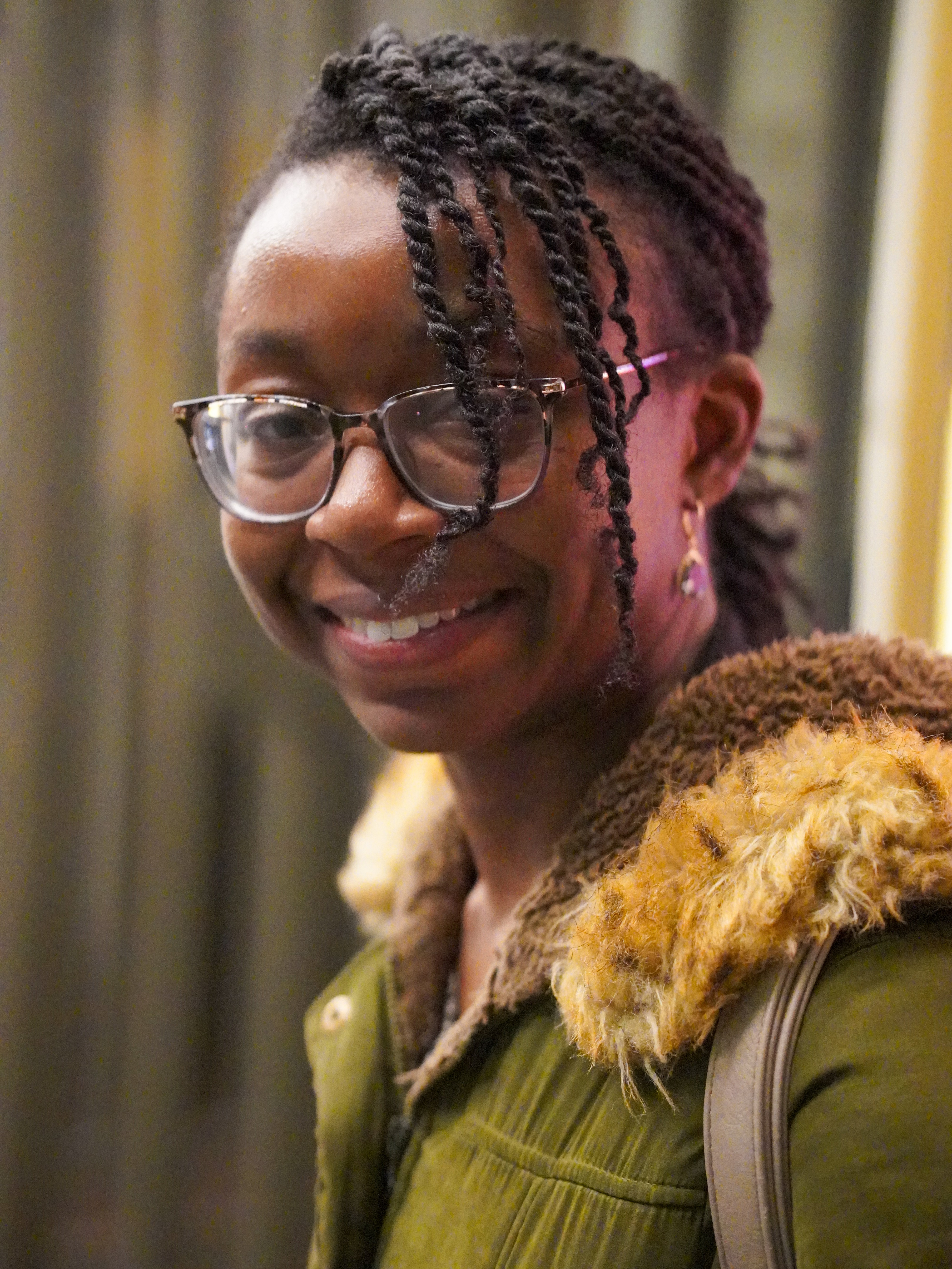}
received her bachelor's degree in Mechanical Engineering from Northeastern University and is currently a Rotational Development Engineer at Boston Dynamics. Her prior experience includes post market engineering analysis of medical devices and kitchen automation robotics. Her engineering interests include robotics, mechanical design, and root cause analysis.
\end{biographywithpic}
\begin{biographywithpic}
{Chloe Wilson}{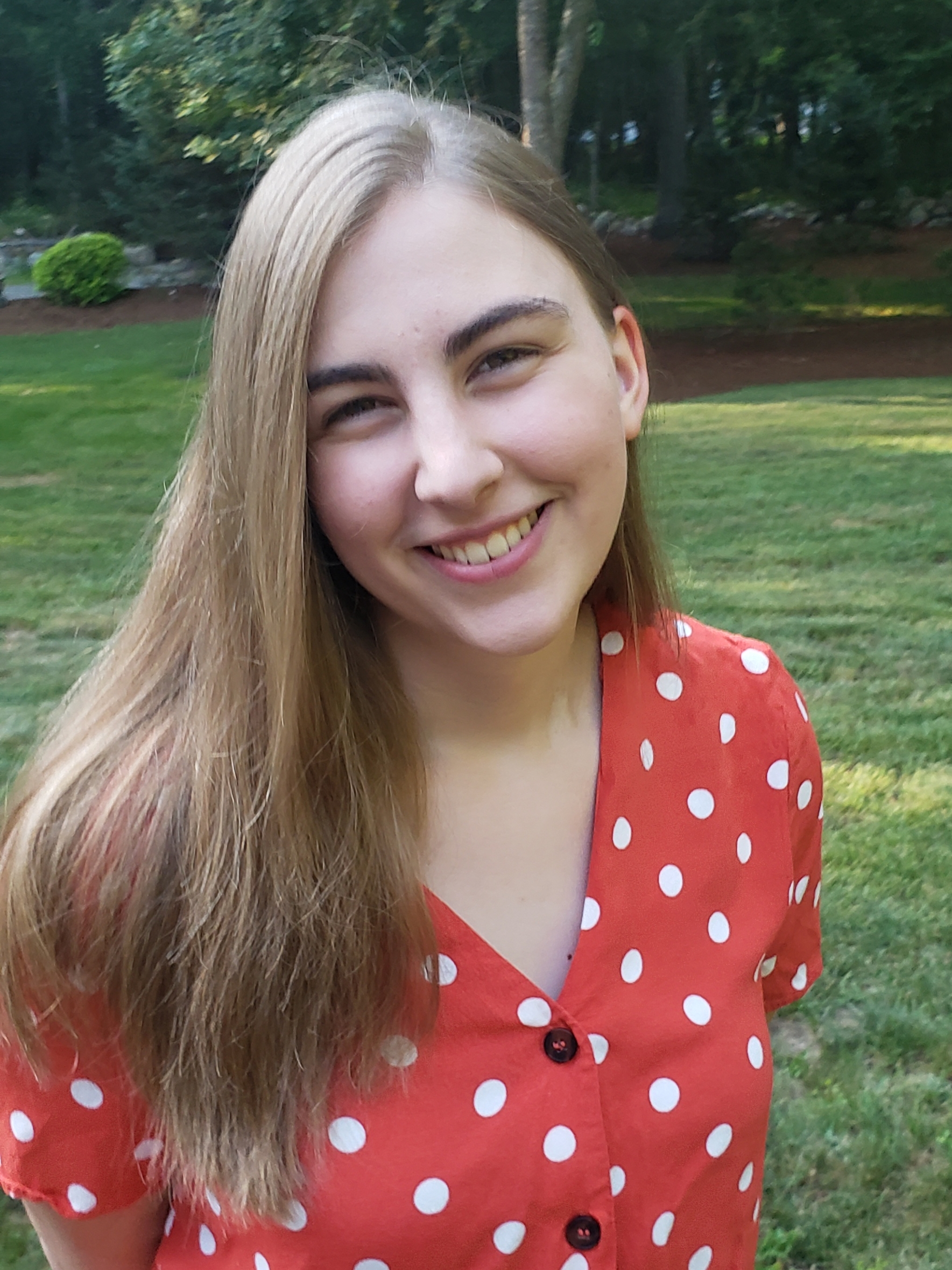}
is an Associate Mechanical Engineer at Nyobolt Inc, a UK-based startup developing fast-charging battery solutions for EVs, industrial robotics, and automation. She received her B.S. in Mechanical Engineering with a minor in Physics from Northeastern University in 2023 where she collaborated on this project.
\end{biographywithpic}
\begin{biographywithpic}
{Bahram Shafai}{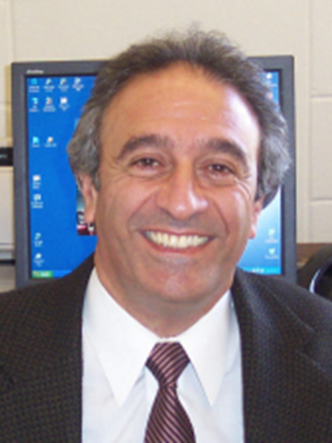}
received his BS and MS in EE from the Swiss Federal Institute of Technology (ETH) Zurich, Switzerland, in 1976 and 1978, respectively, and the Ph.D. degree from GWU in 1985. He joined ECE Department at Northeastern University in 1985. Currently he is a Full Professor and the director of the ECE Capstone Design Program. He is active within IEEE Control Systems Society and published numerous papers mainly in robust stability, observers, and robust control of dynamic systems.
\end{biographywithpic}
\begin{biographywithpic}
{Nathaniel Hanson}{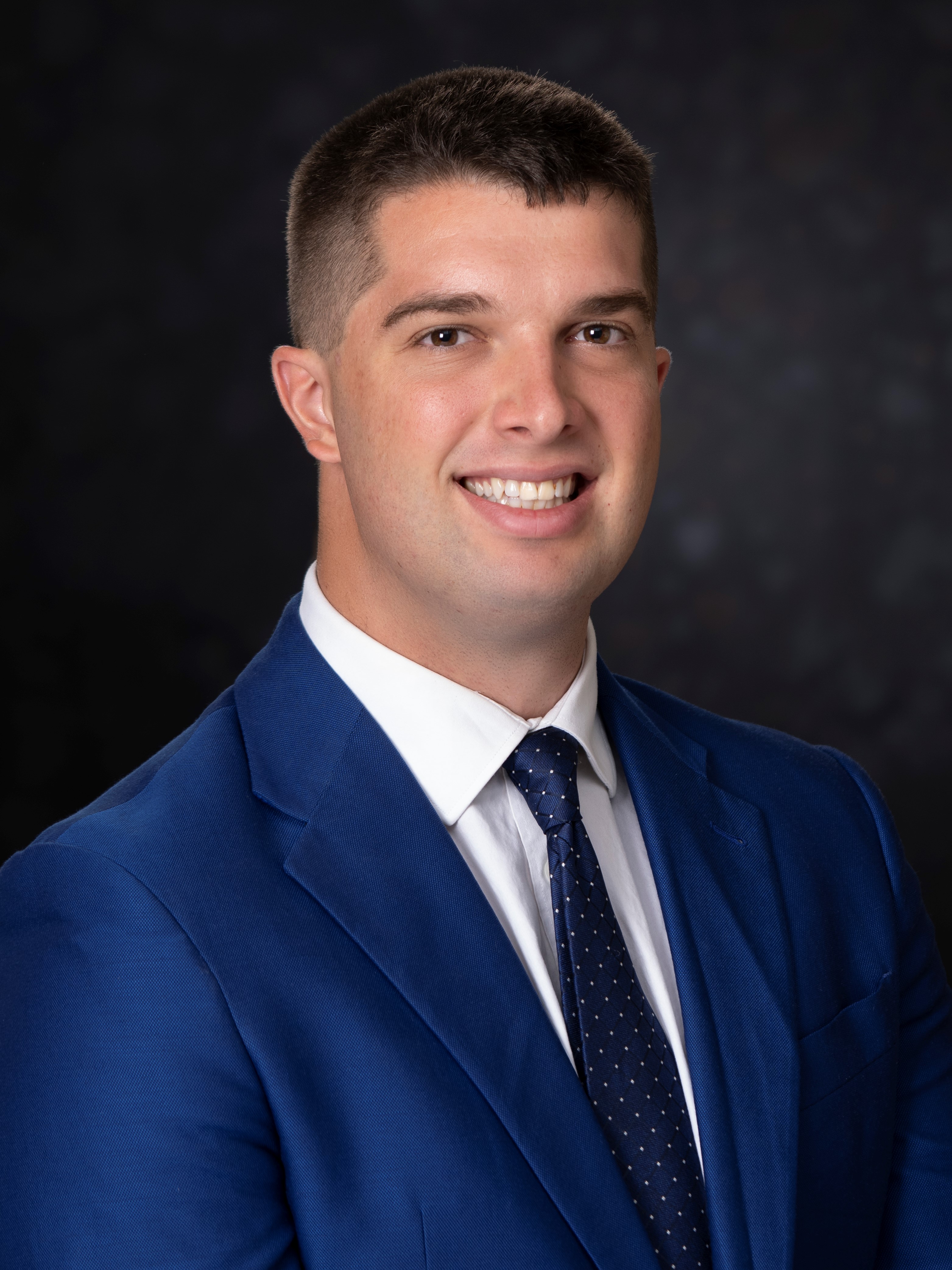}
received his B.S. in Computer Engineering from the University of Notre Dame and M.S. in Computer Science from Boston University, and Ph.D. degree in Computer Engineering from Northeastern University. He is a member of the technical staff at MIT Lincoln Laboratory in the Humanitarian Assistance and Disaster Relief Systems group. His research interests include robot-centric hyperspectral imaging, material recognition, and multi-sensor fusion.
\end{biographywithpic}
\begin{biographywithpic}
{Taşkın Padır}{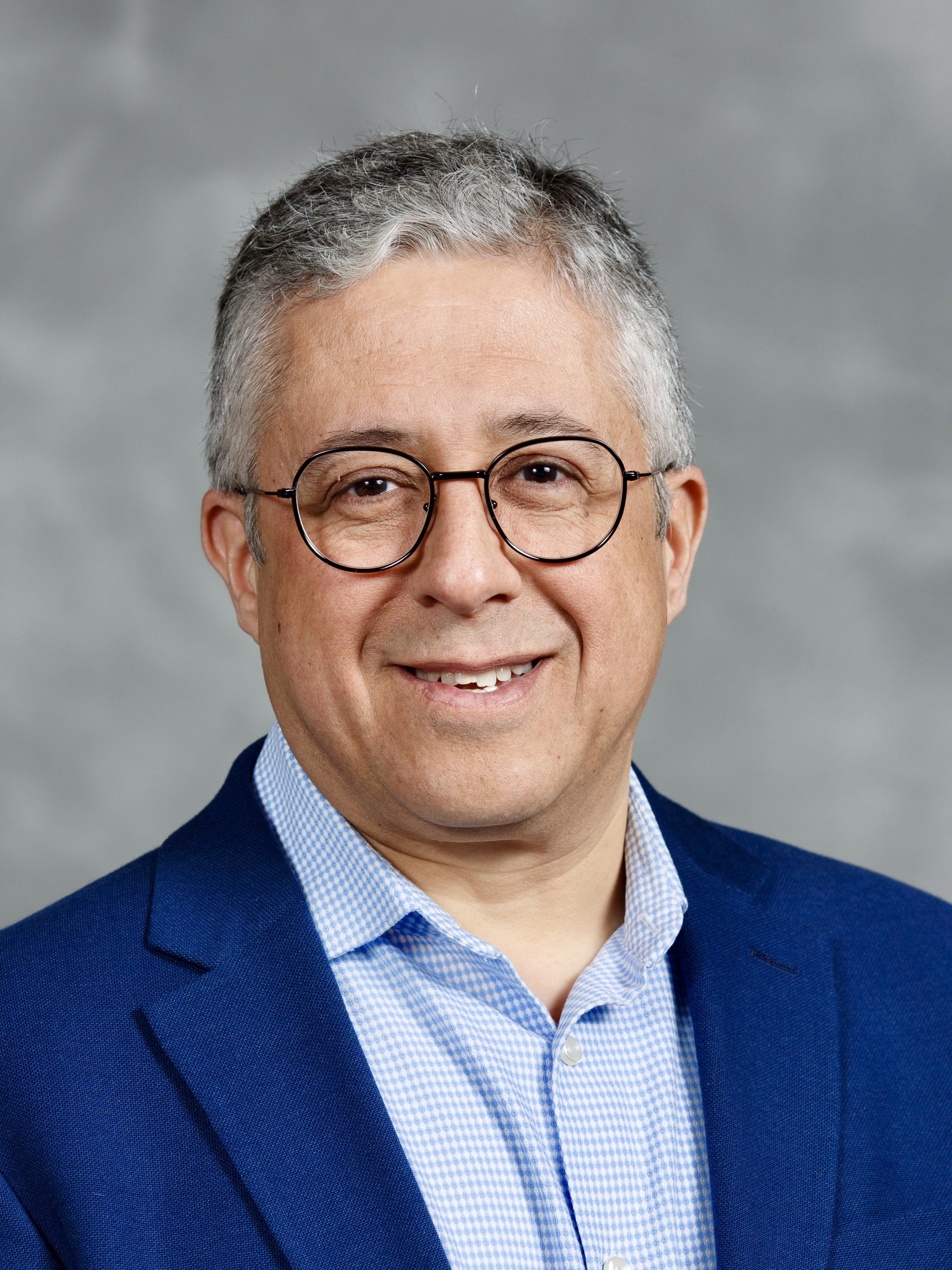}
holds concurrent positions as a Professor in the Electrical and Computer Engineering Department at Northeastern University, and an Amazon Scholar. He received his PhD and MS degrees from Purdue University. He is the Founding Director of the Institute for Experiential Robotics at Northeastern University. His research focuses on shared autonomy, human-in-the-loop robotics, embodied AI, and human-robot teaming at the extremes.
\end{biographywithpic}
\end{document}